\documentclass[article, nojss]{jss}

\usepackage[breakable]{tcolorbox}
\usepackage{parskip} 

\definecolor{urlcolor}{rgb}{0,.145,.698}
\definecolor{linkcolor}{rgb}{.71,0.21,0.01}
\definecolor{citecolor}{rgb}{.12,.54,.11}

\definecolor{ansi-black}{HTML}{3E424D}
\definecolor{ansi-black-intense}{HTML}{282C36}
\definecolor{ansi-red}{HTML}{E75C58}
\definecolor{ansi-red-intense}{HTML}{B22B31}
\definecolor{ansi-green}{HTML}{00A250}
\definecolor{ansi-green-intense}{HTML}{007427}
\definecolor{ansi-yellow}{HTML}{DDB62B}
\definecolor{ansi-yellow-intense}{HTML}{B27D12}
\definecolor{ansi-blue}{HTML}{208FFB}
\definecolor{ansi-blue-intense}{HTML}{0065CA}
\definecolor{ansi-magenta}{HTML}{D160C4}
\definecolor{ansi-magenta-intense}{HTML}{A03196}
\definecolor{ansi-cyan}{HTML}{60C6C8}
\definecolor{ansi-cyan-intense}{HTML}{258F8F}
\definecolor{ansi-white}{HTML}{C5C1B4}
\definecolor{ansi-white-intense}{HTML}{A1A6B2}
\definecolor{ansi-default-inverse-fg}{HTML}{FFFFFF}
\definecolor{ansi-default-inverse-bg}{HTML}{000000}

\DefineVerbatimEnvironment{Highlighting}{Verbatim}{commandchars=\\\{\}}

\let\Oldtex\TeX
\let\Oldlatex\LaTeX
\renewcommand{\TeX}{\textrm{\Oldtex}}
\renewcommand{\LaTeX}{\textrm{\Oldlatex}}
\title{submission\_script}

\makeatletter
\def\PY@reset{\let\PY@it=\relax \let\PY@bf=\relax%
    \let\PY@ul=\relax \let\PY@tc=\relax%
    \let\PY@bc=\relax \let\PY@ff=\relax}
\def\PY@tok#1{\csname PY@tok@#1\endcsname}
\def\PY@toks#1+{\ifx\relax#1\empty\else%
    \PY@tok{#1}\expandafter\PY@toks\fi}
\def\PY@do#1{\PY@bc{\PY@tc{\PY@ul{%
    \PY@it{\PY@bf{\PY@ff{#1}}}}}}}
\def\PY#1#2{\PY@reset\PY@toks#1+\relax+\PY@do{#2}}

\@namedef{PY@tok@w}{\def\PY@tc##1{\textcolor[rgb]{0.73,0.73,0.73}{##1}}}
\@namedef{PY@tok@c}{\let\PY@it=\textit\def\PY@tc##1{\textcolor[rgb]{0.24,0.48,0.48}{##1}}}
\@namedef{PY@tok@cp}{\def\PY@tc##1{\textcolor[rgb]{0.61,0.40,0.00}{##1}}}
\@namedef{PY@tok@k}{\let\PY@bf=\textbf\def\PY@tc##1{\textcolor[rgb]{0.00,0.50,0.00}{##1}}}
\@namedef{PY@tok@kp}{\def\PY@tc##1{\textcolor[rgb]{0.00,0.50,0.00}{##1}}}
\@namedef{PY@tok@kt}{\def\PY@tc##1{\textcolor[rgb]{0.69,0.00,0.25}{##1}}}
\@namedef{PY@tok@o}{\def\PY@tc##1{\textcolor[rgb]{0.40,0.40,0.40}{##1}}}
\@namedef{PY@tok@ow}{\let\PY@bf=\textbf\def\PY@tc##1{\textcolor[rgb]{0.67,0.13,1.00}{##1}}}
\@namedef{PY@tok@nb}{\def\PY@tc##1{\textcolor[rgb]{0.00,0.50,0.00}{##1}}}
\@namedef{PY@tok@nf}{\def\PY@tc##1{\textcolor[rgb]{0.00,0.00,1.00}{##1}}}
\@namedef{PY@tok@nc}{\let\PY@bf=\textbf\def\PY@tc##1{\textcolor[rgb]{0.00,0.00,1.00}{##1}}}
\@namedef{PY@tok@nn}{\let\PY@bf=\textbf\def\PY@tc##1{\textcolor[rgb]{0.00,0.00,1.00}{##1}}}
\@namedef{PY@tok@ne}{\let\PY@bf=\textbf\def\PY@tc##1{\textcolor[rgb]{0.80,0.25,0.22}{##1}}}
\@namedef{PY@tok@nv}{\def\PY@tc##1{\textcolor[rgb]{0.10,0.09,0.49}{##1}}}
\@namedef{PY@tok@no}{\def\PY@tc##1{\textcolor[rgb]{0.53,0.00,0.00}{##1}}}
\@namedef{PY@tok@nl}{\def\PY@tc##1{\textcolor[rgb]{0.46,0.46,0.00}{##1}}}
\@namedef{PY@tok@ni}{\let\PY@bf=\textbf\def\PY@tc##1{\textcolor[rgb]{0.44,0.44,0.44}{##1}}}
\@namedef{PY@tok@na}{\def\PY@tc##1{\textcolor[rgb]{0.41,0.47,0.13}{##1}}}
\@namedef{PY@tok@nt}{\let\PY@bf=\textbf\def\PY@tc##1{\textcolor[rgb]{0.00,0.50,0.00}{##1}}}
\@namedef{PY@tok@nd}{\def\PY@tc##1{\textcolor[rgb]{0.67,0.13,1.00}{##1}}}
\@namedef{PY@tok@s}{\def\PY@tc##1{\textcolor[rgb]{0.73,0.13,0.13}{##1}}}
\@namedef{PY@tok@sd}{\let\PY@it=\textit\def\PY@tc##1{\textcolor[rgb]{0.73,0.13,0.13}{##1}}}
\@namedef{PY@tok@si}{\let\PY@bf=\textbf\def\PY@tc##1{\textcolor[rgb]{0.64,0.35,0.47}{##1}}}
\@namedef{PY@tok@se}{\let\PY@bf=\textbf\def\PY@tc##1{\textcolor[rgb]{0.67,0.36,0.12}{##1}}}
\@namedef{PY@tok@sr}{\def\PY@tc##1{\textcolor[rgb]{0.64,0.35,0.47}{##1}}}
\@namedef{PY@tok@ss}{\def\PY@tc##1{\textcolor[rgb]{0.10,0.09,0.49}{##1}}}
\@namedef{PY@tok@sx}{\def\PY@tc##1{\textcolor[rgb]{0.00,0.50,0.00}{##1}}}
\@namedef{PY@tok@m}{\def\PY@tc##1{\textcolor[rgb]{0.40,0.40,0.40}{##1}}}
\@namedef{PY@tok@gh}{\let\PY@bf=\textbf\def\PY@tc##1{\textcolor[rgb]{0.00,0.00,0.50}{##1}}}
\@namedef{PY@tok@gu}{\let\PY@bf=\textbf\def\PY@tc##1{\textcolor[rgb]{0.50,0.00,0.50}{##1}}}
\@namedef{PY@tok@gd}{\def\PY@tc##1{\textcolor[rgb]{0.63,0.00,0.00}{##1}}}
\@namedef{PY@tok@gi}{\def\PY@tc##1{\textcolor[rgb]{0.00,0.52,0.00}{##1}}}
\@namedef{PY@tok@gr}{\def\PY@tc##1{\textcolor[rgb]{0.89,0.00,0.00}{##1}}}
\@namedef{PY@tok@ge}{\let\PY@it=\textit}
\@namedef{PY@tok@gs}{\let\PY@bf=\textbf}
\@namedef{PY@tok@ges}{\let\PY@bf=\textbf\let\PY@it=\textit}
\@namedef{PY@tok@gp}{\let\PY@bf=\textbf\def\PY@tc##1{\textcolor[rgb]{0.00,0.00,0.50}{##1}}}
\@namedef{PY@tok@go}{\def\PY@tc##1{\textcolor[rgb]{0.44,0.44,0.44}{##1}}}
\@namedef{PY@tok@gt}{\def\PY@tc##1{\textcolor[rgb]{0.00,0.27,0.87}{##1}}}
\@namedef{PY@tok@err}{\def\PY@bc##1{{\setlength{\fboxsep}{\string -\fboxrule}\fcolorbox[rgb]{1.00,0.00,0.00}{1,1,1}{\strut ##1}}}}
\@namedef{PY@tok@kc}{\let\PY@bf=\textbf\def\PY@tc##1{\textcolor[rgb]{0.00,0.50,0.00}{##1}}}
\@namedef{PY@tok@kd}{\let\PY@bf=\textbf\def\PY@tc##1{\textcolor[rgb]{0.00,0.50,0.00}{##1}}}
\@namedef{PY@tok@kn}{\let\PY@bf=\textbf\def\PY@tc##1{\textcolor[rgb]{0.00,0.50,0.00}{##1}}}
\@namedef{PY@tok@kr}{\let\PY@bf=\textbf\def\PY@tc##1{\textcolor[rgb]{0.00,0.50,0.00}{##1}}}
\@namedef{PY@tok@bp}{\def\PY@tc##1{\textcolor[rgb]{0.00,0.50,0.00}{##1}}}
\@namedef{PY@tok@fm}{\def\PY@tc##1{\textcolor[rgb]{0.00,0.00,1.00}{##1}}}
\@namedef{PY@tok@vc}{\def\PY@tc##1{\textcolor[rgb]{0.10,0.09,0.49}{##1}}}
\@namedef{PY@tok@vg}{\def\PY@tc##1{\textcolor[rgb]{0.10,0.09,0.49}{##1}}}
\@namedef{PY@tok@vi}{\def\PY@tc##1{\textcolor[rgb]{0.10,0.09,0.49}{##1}}}
\@namedef{PY@tok@vm}{\def\PY@tc##1{\textcolor[rgb]{0.10,0.09,0.49}{##1}}}
\@namedef{PY@tok@sa}{\def\PY@tc##1{\textcolor[rgb]{0.73,0.13,0.13}{##1}}}
\@namedef{PY@tok@sb}{\def\PY@tc##1{\textcolor[rgb]{0.73,0.13,0.13}{##1}}}
\@namedef{PY@tok@sc}{\def\PY@tc##1{\textcolor[rgb]{0.73,0.13,0.13}{##1}}}
\@namedef{PY@tok@dl}{\def\PY@tc##1{\textcolor[rgb]{0.73,0.13,0.13}{##1}}}
\@namedef{PY@tok@s2}{\def\PY@tc##1{\textcolor[rgb]{0.73,0.13,0.13}{##1}}}
\@namedef{PY@tok@sh}{\def\PY@tc##1{\textcolor[rgb]{0.73,0.13,0.13}{##1}}}
\@namedef{PY@tok@s1}{\def\PY@tc##1{\textcolor[rgb]{0.73,0.13,0.13}{##1}}}
\@namedef{PY@tok@mb}{\def\PY@tc##1{\textcolor[rgb]{0.40,0.40,0.40}{##1}}}
\@namedef{PY@tok@mf}{\def\PY@tc##1{\textcolor[rgb]{0.40,0.40,0.40}{##1}}}
\@namedef{PY@tok@mh}{\def\PY@tc##1{\textcolor[rgb]{0.40,0.40,0.40}{##1}}}
\@namedef{PY@tok@mi}{\def\PY@tc##1{\textcolor[rgb]{0.40,0.40,0.40}{##1}}}
\@namedef{PY@tok@il}{\def\PY@tc##1{\textcolor[rgb]{0.40,0.40,0.40}{##1}}}
\@namedef{PY@tok@mo}{\def\PY@tc##1{\textcolor[rgb]{0.40,0.40,0.40}{##1}}}
\@namedef{PY@tok@ch}{\let\PY@it=\textit\def\PY@tc##1{\textcolor[rgb]{0.24,0.48,0.48}{##1}}}
\@namedef{PY@tok@cm}{\let\PY@it=\textit\def\PY@tc##1{\textcolor[rgb]{0.24,0.48,0.48}{##1}}}
\@namedef{PY@tok@cpf}{\let\PY@it=\textit\def\PY@tc##1{\textcolor[rgb]{0.24,0.48,0.48}{##1}}}
\@namedef{PY@tok@c1}{\let\PY@it=\textit\def\PY@tc##1{\textcolor[rgb]{0.24,0.48,0.48}{##1}}}
\@namedef{PY@tok@cs}{\let\PY@it=\textit\def\PY@tc##1{\textcolor[rgb]{0.24,0.48,0.48}{##1}}}

\def\PYZbs{\char`\\}
\def\PYZus{\char`\_}
\def\PYZob{\char`\{}
\def\PYZcb{\char`\}}

\def\PYZpc{\char`\%}

\def\PYZhy{\char`\-}
\def\PYZsq{\char`\'}
\def\PYZdq{\char`\"}

\makeatother

\makeatletter
    \newbox\Wrappedcontinuationbox
    \newbox\Wrappedvisiblespacebox
    \newcommand*\Wrappedvisiblespace {\textcolor{red}{\textvisiblespace}}
    \newcommand*\Wrappedcontinuationsymbol {\textcolor{red}{\llap{\tiny$\m@th\hookrightarrow$}}}
    \newcommand*\Wrappedcontinuationindent {3ex }
    \newcommand*\Wrappedafterbreak {\kern\Wrappedcontinuationindent\copy\Wrappedcontinuationbox}
    \newcommand*\Wrappedbreaksatspecials {%
        \def\PYGZus{\discretionary{\char`\_}{\Wrappedafterbreak}{\char`\_}}%
        \def\PYGZob{\discretionary{}{\Wrappedafterbreak\char`\{}{\char`\{}}%
        \def\PYGZcb{\discretionary{\char`\}}{\Wrappedafterbreak}{\char`\}}}%
        \def\PYGZca{\discretionary{\char`\^}{\Wrappedafterbreak}{\char`\^}}%
        \def\PYGZam{\discretionary{\char`\&}{\Wrappedafterbreak}{\char`\&}}%
        \def\PYGZlt{\discretionary{}{\Wrappedafterbreak\char`\<}{\char`\<}}%
        \def\PYGZgt{\discretionary{\char`\>}{\Wrappedafterbreak}{\char`\>}}%
        \def\PYGZsh{\discretionary{}{\Wrappedafterbreak\char`\#}{\char`\#}}%
        \def\PYGZpc{\discretionary{}{\Wrappedafterbreak\char`\%}{\char`\%}}%
        \def\PYGZdl{\discretionary{}{\Wrappedafterbreak\char`\$}{\char`\$}}%
        \def\PYGZhy{\discretionary{\char`\-}{\Wrappedafterbreak}{\char`\-}}%
        \def\PYGZsq{\discretionary{}{\Wrappedafterbreak\textquotesingle}{\textquotesingle}}%
        \def\PYGZdq{\discretionary{}{\Wrappedafterbreak\char`\"}{\char`\"}}%
        \def\PYGZti{\discretionary{\char`\~}{\Wrappedafterbreak}{\char`\~}}%
    }
    \newcommand*\Wrappedbreaksatpunct {%
        \lccode`\~`\.\lowercase{\def~}{\discretionary{\hbox{\char`\.}}{\Wrappedafterbreak}{\hbox{\char`\.}}}%
        \lccode`\~`\,\lowercase{\def~}{\discretionary{\hbox{\char`\,}}{\Wrappedafterbreak}{\hbox{\char`\,}}}%
        \lccode`\~`\;\lowercase{\def~}{\discretionary{\hbox{\char`\;}}{\Wrappedafterbreak}{\hbox{\char`\;}}}%
        \lccode`\~`\:\lowercase{\def~}{\discretionary{\hbox{\char`\:}}{\Wrappedafterbreak}{\hbox{\char`\:}}}%
        \lccode`\~`\?\lowercase{\def~}{\discretionary{\hbox{\char`\?}}{\Wrappedafterbreak}{\hbox{\char`\?}}}%
        \lccode`\~`\!\lowercase{\def~}{\discretionary{\hbox{\char`\!}}{\Wrappedafterbreak}{\hbox{\char`\!}}}%
        \lccode`\~`\/\lowercase{\def~}{\discretionary{\hbox{\char`\/}}{\Wrappedafterbreak}{\hbox{\char`\/}}}%
        \catcode`\.\active
        \catcode`\,\active
        \catcode`\;\active
        \catcode`\:\active
        \catcode`\?\active
        \catcode`\!\active
        \catcode`\/\active
        \lccode`\~`\~
    }
\makeatother

\let\OriginalVerbatim=\Verbatim
\makeatletter
\renewcommand{\Verbatim}[1][1]{%
    \sbox\Wrappedcontinuationbox {\Wrappedcontinuationsymbol}%
    \sbox\Wrappedvisiblespacebox {\FV@SetupFont\Wrappedvisiblespace}%
    \def\FancyVerbFormatLine ##1{\hsize\linewidth
        \vtop{\raggedright\hyphenpenalty\z@\exhyphenpenalty\z@
            \doublehyphendemerits\z@\finalhyphendemerits\z@
            \strut ##1\strut}%
    }%
    \def\FV@Space {%
        \nobreak\hskip\z@ plus\fontdimen3\font minus\fontdimen4\font
        \discretionary{\copy\Wrappedvisiblespacebox}{\Wrappedafterbreak}
        {\kern\fontdimen2\font}%
    }%

    \Wrappedbreaksatspecials
    \OriginalVerbatim[#1,codes*=\Wrappedbreaksatpunct]%
}
\makeatother

\definecolor{incolor}{HTML}{303F9F}
\definecolor{outcolor}{HTML}{D84315}
\definecolor{cellborder}{HTML}{CFCFCF}
\definecolor{cellbackground}{HTML}{F7F7F7}

\makeatletter
\newcommand{\boxspacing}{\kern\kvtcb@left@rule\kern\kvtcb@boxsep}
\makeatother

\hypersetup{
  breaklinks=true,  
  colorlinks=true,
  urlcolor=urlcolor,
  linkcolor=linkcolor,
  citecolor=citecolor,
  }

\usepackage{orcidlink,thumbpdf,lmodern}

\usepackage{framed}

\usepackage{breqn}

\usepackage{enumitem}
\setlist{itemsep=0.2ex, topsep=0ex, partopsep=0ex, parsep=0.5ex, leftmargin=2.3ex}
\usepackage{bm, bbm}
\def\indep{\perp \!\!\! \perp}
\def\notindep{\not\!\perp\!\!\!\perp}
\usepackage{pifont} 
\usepackage{xcolor} 
\usepackage{booktabs}
\usepackage{amssymb}
\usepackage{bm, bbm}
\definecolor{lessbrightgreen}{rgb}{0.0, 0.5, 0.0} 
\newcommand{\cmark}{\textcolor{lessbrightgreen}{\ding{51}}} 
\newcommand{\xmark}{\textcolor{red!70!black}{\ding{55}}}

\definecolor{purple}{rgb}{0.6, 0.0, 0.6} 

\author{Mike Van Ness \\Stanford University \And Madeleine Udell \\Stanford University}
\Plainauthor{Mike Van Ness, Madeleine Udell}

\title{dnamite: A Python Package for \\ Neural Additive Models}
\Plaintitle{dnamite: A Python Package for Neural Additive Models}
\Shorttitle{dnamite Package}

\Abstract{
Additive models offer accurate and interpretable predictions for tabular data,
a critical tool for statistical modeling.
Recent advances in Neural Additive Models (NAMs) allow these models to handle complex machine learning tasks, including feature selection and survival analysis, 
on large-scale data. 
This paper introduces \textbf{dnamite}, a Python package
that implements NAMs for these advanced applications.
\textbf{dnamite} provides a scikit-learn style interface to train regression, classification, and survival analysis NAMs, 
with built-in support for feature selection. 
We describe the methodology underlying \textbf{dnamite}, its design principles, and its implementation. 
Through an application to the MIMIC III clinical dataset, we demonstrate the utility of \textbf{dnamite} in a real-world setting where feature selection and survival analysis are both important. The package is publicly available via pip and documented at \url{dnamite.readthedocs.io}.
}

\Plainkeywords{additive models, feature selection, survival analysis, interpretable machine learning}

\begin{document}

\section{Introduction} \label{sec:intro}

Additive models are an important statistical tool for predictive modeling
\citep{hastie2017generalized, wood2017generalized}.
As in linear models, 
the final prediction of an additive model 
is the sum of a prediction for each feature.
But whereas a linear model uses a scalar multiple of each feature as its contribution,
an additive model uses a \textit{nonlinear} model.
This nonlinearity allows additive models to achieve better predictive accuracy than linear models while maintaining some of the structure and interpretability of linear models.

Modern machine learning practice favors model classes such as gradient-boosted decision trees \citep{anghel2018benchmarking} and neural networks \citep{borisov2022deep}
for their excellent predictive performance.
However, most of these models are \emph{black-box models}, meaning it is difficult to understand why they make a particular prediction 
without (or even with) extra post-hoc analysis.
Practitioners in domains like healthcare are reasonably hesitant to use such black-box models, as these models are difficult to understand fully and can obscure harmful perversities in the data \citep{caruana2015intelligible, bolukbasi2016man}.
Hence, additive models, rebranded as interpretable machine learning models \citep{nori2019interpretml, sudjianto2023piml}, have seen new popularity as a more explainable alternative to black-box models.
This new generation of additive models employs modern machine learning techniques such as gradient-boosted trees, neural networks, and hardware accelerators,
delivering models that compare well with state-of-the-art black-box models on tabular data.

Most recently, additive models trained with neural networks, Neural Additive Models (NAMs), have gained popularity 
as a means to adapt additive models to more advanced applications 
\citep{agarwal2021neural, chang2021node, yang2021gami},
including feature selection
for high-dimensional problems 
\citep{yang2021gami, ibrahim2024grand, xu2023sparse},
and survival analysis
\citep{rahman2021pseudonam, kopper2022deeppamm, van2023interpretable, van2024interpretable, van2024dnamite}.
While standard additive models have support through mature packages in R \citep{wood2001mgcv} and Python \citep{serven2018pygam}, 
there is currently no package in R or Python for NAMs that supports these more advanced applications.

To fill this gap, we introduce \textbf{dnamite}, a Python package for neural additive modeling. 
\textbf{dnamite} implements NAMs via the DNAMite architecture \citep{van2024dnamite}, which was originally proposed for survival analysis but can be easily adapted to support regression and classification.
Additionally, \textbf{dnamite} supports feature selection for regression, classification, and survival analysis using the learnable gates method \citep{ibrahim2024grand, van2024interpretable}. 
Most importantly, \textbf{dnamite} implements the scikit-learn interface, making training and analysis accessible with simple function calls.
\textbf{dnamite} is open-source, publicly available on github, registered for installation via pip, and fully documented at \url{https://dnamite.readthedocs.io}. 

This paper presents the methods used by \textbf{dnamite} and showcases both standard and advanced usage of the package. Section \ref{sec:background} reviews \textbf{dnamite}'s mathematical foundations and compares \textbf{dnamite} with other additive model packages. Section \ref{sec:methodology} introduces the DNAMite architecture, with applications to feature selection and survival analysis. Section \ref{sec:design} outlines the principles that motivate our design of the \textbf{dnamite} package. 
Section \ref{usage} demonstrates the practical usage of \textbf{dnamite} with the MIMIC III clinical dataset. Finally, Section \ref{sec:conclusion} offers a summary and concluding remarks.

\section{Background} \label{sec:background}

\textbf{dnamite} fits additive models to tabular datasets: data of the form $(X, Y)$ where $X = (X_1, X_2, \ldots, X_p)$ is a vector of $p$ features and $Y$ is a supervised label.
Each feature $X_j$ can be continuous or categorical and can be missing.
For regression $Y \in \mathbb{R}$, for binary classification $Y \in \{0, 1\}$, and for survival analysis $Y = (Z, \delta)$ where $Z \in \mathbb{R}^+$ is a time-to-event label and $\delta \in \{0, 1\}$ is a censor indicator (see Section \ref{sec:survival_analysis}).
\textbf{dnamite} aims to train models $f(X)$ that predict $Y$ accurately while being explainable to the user by design.

\subsection{Additive Models}

An additive model $f(X)$ with task-dependent link function $g$ takes the form
\begin{equation}
    f(X) = g\big( f_1(X_1) + f_2(X_2) + \cdots + f_p(X_p) \big).
\end{equation}
Additive models generalize linear models by allowing each feature function \( f_j \), known as a \textit{shape function}, to be nonlinear, thereby capturing arbitrarily complex patterns.
Additive models enjoy many of the benefits of linear models but can provide significantly more accuracy.
First, because additive models preserve additive structure, plotting $\hat{f}_j$ visualizes the exact contribution of $X_j$ to the final (unlinked) prediction.
Second, averaging $|\hat{f}_j|$ across the training data gives a natural importance score for feature $j$, 
allowing for easy assessment of the most important features.

Additive models can also support feature interactions: 
given a set of interactions $\mathcal{I} \subset \{i \neq j \mid 1 \leq i,j \leq p \}$, 
an additive model with interactions takes the form
\begin{equation}
    f(X) = g \left( \sum_{j=1}^p f_j(X_j) + \sum_{i, j \in \mathcal{I}} f_{i, j}(X_i, X_j) \right).
\end{equation}
We call the single-feature shape functions \textit{main effects} and the pairwise interactions \textit{interaction effects},
extending the terminology used in linear models.

Interaction effects can improve predictive performance but risk making model predictions less interpretable.
Specifically, if feature $j$ appears in at least one interaction term, 
then the shape function $\hat{f}_j$ no longer represents the entire contribution of $X_j$ to the final prediction.
Most additive models with interactions take one of two approaches to mitigate this problem.
The first approach fits and freezes all main effects first before fitting interaction effects 
so that the main effects are the same regardless of whether or not interactions are included.
The second approach jointly fits mains and interactions and then uses a purification process 
to ensure any signal that can be explained by an individual feature alone appears in a main effect and not an interaction \citep{lengerich2020purifying}.

The literature offers several model classes to represent shape functions.
The \textbf{dnamite} package uses Neural Additive Models (NAMs), additive models that use neural networks to represent each shape function $f_j$.
NAMs harness the flexibility and scalability of modern neural network design and optimization to adapt to complex problems more easily than other model classes.
\textbf{dnamite} exploits this flexibility to offer feature selection and survival analysis using NAMs.

\subsection{Survival Analysis}
\label{sec:survival_analysis}

Survival analysis is a classic problem in statistics that models time-to-event data.
Survival analysis is widely used in healthcare statistics \citep{lee1997survival},
but is also used in many domains such as epidemiology \citep{selvin2008survival} and economics \citep{danacica2010using}.
Improving the quality of additive models for survival analysis is an important challenge
as both accuracy and interpretability are important in these applications.

In survival analysis, the goal is to estimate the conditional survival curve $P(T > t \mid X)$ for a time-to-event label $T > 0$.
However, $T$ is assumed to be unavailable for some samples.
Instead, we only know that no event occurred for these samples up until a censoring time $C > 0$.
Thus, we observe the label $(Z, \delta)$, where $Z = \min(C, T)$ is the observed event or censoring time and $\delta = \mathbbm{1}_{C > T}$ is the censoring indicator.

One approach to estimating $P(T > t \mid X)$ is to first estimate the conditional hazard function
\begin{equation}
    \lambda(t \mid X) = \frac{f_{T \mid X}(t)}{P(T > t \mid X)}
\end{equation}
and then use the relationship
\begin{equation}
    P(T > t \mid X) = \exp \left( - \int_0^t \lambda(s \mid X) ds \right).
\end{equation}
The most classical approach is the Cox proportional hazards model \citep{cox1972regression}, which assumes a linear form for the logarithmic conditional hazard:
\begin{equation}
    \lambda(t \mid X) = \lambda_0(t) \exp(\beta_1 X_1 + \cdots + \beta_p X_p).
\end{equation}
The benefit of the Cox model is that inference on the regression coefficients does not require 
estimating $\lambda_0(t)$, which greatly simplifies the problem.
On the other hand, the Cox model (or any nonlinear generalization) 
assumes that the hazard function is proportional over time, as
\begin{equation}
    \frac{\lambda(t \mid X^{(1)})}{\lambda(t \mid X^{(2)})} = \frac{\exp(\beta_1 X^{(1)}_1 + \cdots + \beta_p X^{(1)}_p)}{\exp(\beta_1 X^{(2)}_1 + \cdots + \beta_p X^{(2)}_p)}, \quad \forall t.
\end{equation}
More recently, several deep learning models have been introduced that can directly estimate the conditional hazard function or the conditional survival curve without simplifying assumptions \citep{chen2024introduction}.
\textbf{dnamite} estimates the conditional survival curve in this way.

\subsection{Related Packages}
\label{sec:related_packages}

Several existing packages can be used to train additive models, see Table \ref{tab:related_work}.
In R, the original gam package \citep{hastie2024gam} as well as its successor mgcv \citep{wood2001mgcv} are mature packages for fitting additive models.
Both packages use splines or other scatterplot smoothers to estimate shape functions.
In Python, PyGAM \citep{serven2018pygam} uses splines while interpretml \citep{seabold2010statsmodels} uses boosted decision trees to fit additive models.
However, neither of these packages support feature selection.
Additionally, PyGAM and interpretml do not support survival analysis, while gam and mgcv only support survival models with proportional hazards.

Surprisingly, very few packages exist for training NAMs. 
The PiML package \citep{sudjianto2023piml} implements several NAM architectures, but only one (GAMI-Net \citep{yang2021gami}) that supports feature-sparsity and none that support survival analysis.
The R package neuralGAM \citep{ortega2024explainable} also has a Python version but with no support for feature selection or survival analysis.
The \textbf{dnamite} package, meanwhile, supports feature selection and survival analysis.
Moreover, \textbf{dnamite} implements a NAM architecture that has been shown to fit shape functions more accurately than standard NAM architectures \citep{van2024dnamite}.

\begin{table}[]
    \centering
    \caption{Comparison of additive model packages.}
    \vspace{0.5em}
    \begin{tabular}{|c|c|c|c|c|}
    \toprule
    Package & Language & Interactions & Feature Selection & Survival Analysis \\
    \midrule
        gam & R & \cmark & \xmark & \cmark \\
        mgcv & R & \cmark & \xmark & \cmark \\
        neuralGAM & R/Python & \xmark & \xmark & \xmark \\
        \midrule
        PyGAM & Python & \cmark & \xmark & \xmark  \\
        interpretml & Python & \cmark & interactions only & \xmark \\
        PiML & Python & \cmark & \cmark & \xmark \\
        \midrule
        dnamite & Python & \cmark & \cmark & \cmark \\
        \bottomrule
    \end{tabular}
    \label{tab:related_work}
\end{table}

\section{Methodology} \label{sec:methodology}

\subsection{Model Architecture}
\label{sec:architecture}

\begin{figure}[t]
    \centering
    \includegraphics[width=\linewidth]{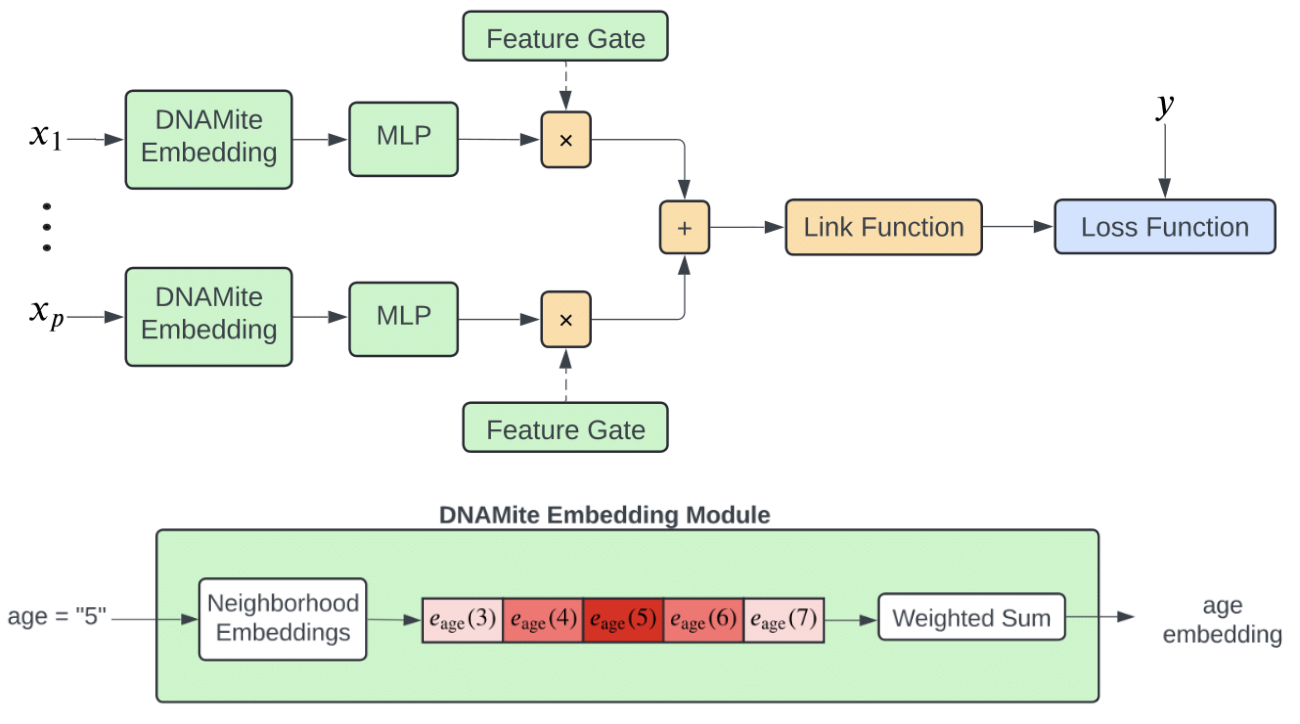}
    \caption{The DNAMite architecture. For simplicity, only individual features (and not interactions) are shown. The choice of link and loss function depends on the prediction task. Feature gates are only used for feature selection. The embedding module outputs a weighted sum of neighborhood embeddings for each feature.}
    \label{fig:architecture}
\end{figure}

\textbf{dnamite} uses the eponymous DNAMite model architecture \citep{van2024dnamite} for all NAM implementations, summarized in Figure \ref{fig:architecture}.
While the DNAMite architecture was designed for survival analysis, it can easily be adapted for regression and classification by changing the output dimension, link function, or loss function.

DNAMite fits additive models with structure
\begin{equation}
\label{eq:model_with_zs}
    f(X) = g\left( \beta_0 + \sum_{j=1}^p s(\mu_j) f_j(X_j) + \sum_{j < k} s(\mu_{j, k}) f_{j, k}(X_j, X_k) \right)
\end{equation}
where the link function $g$ is set based on the task. 
The feature gates $s(\mu_j), s(\mu_{j, k}) \in [0, 1]$ facilitate feature selection, detailed in Section \ref{sec:feature_selection}.
Interactions are fit in a second round of training after freezing the fitted main effects.
The intercept term $\beta_0$ is estimated after training:
estimated main and interaction shape functions are centered
\begin{equation}
\label{eq:centering}
\hat{f}_j - \frac{1}{n}\sum_{i=1}^n s(\hat{\mu}_j) \hat{f}_j(X_j^{(i)}), \quad \hat{f}_{j, k} - \frac{1}{n}\sum_{i=1}^n s(\hat{\mu}_{j, k}) \hat{f}_{j, k}(X_j^{(i)}, X_k^{(i)})
\end{equation}
and the intercept is set to 
\begin{equation}
\label{eq:intercept}
    \hat{\beta}_0 = \sum_{j=1}^p \left( \frac{1}{n} \sum_{i=1}^n s(\hat{\mu}_j) \hat{f}_j (X_j^{(i)})\right) + \sum_{j, k \in \mathcal{I}} \left( \frac{1}{n} \sum_{i=1}^n s(\hat{\mu}_{j, k}) \hat{f}_{j, k} (X_j^{(i)}, X_k^{(i)})\right).
\end{equation}
Centering ensures that shape functions are identifiable \citep{van2024dnamite}.

A key component of the DNAMite architecture is its feature embedding module, see the bottom of Figure \ref{fig:architecture}.
This module first discretizes all features using feature-specific bins.
For categorical features, one bin is given to each unique feature value.
For continuous features, bins are defined using quantiles so that each bin has approximately the same number of samples in the training dataset.
One additional bin per feature encodes missing values.
After discretization, the embedding module learns an embedding for each feature bin.
The module's final embedding, then, is a weighted sum of embeddings in a neighborhood around the input bin.
Mathematically, given an input bin index $i \in \mathbb{Z}^+$, kernel size $K$, and embedding function $e : \mathbb{Z}^+ \to \mathbb{R}^d$, the final embedding for input bin index $i$ is
\begin{equation}
    \text{embed}(i) = \sum_{j = -K}^K w_j e(i + j)
    \qquad \text{where} \qquad 
    w_j = \exp\left( - \frac{(i - j)^2}{2 \phi}\right). \label{eq:gaussian_kernel}
\end{equation}
The parameter $\phi$, which sets the variance of the Gaussian kernel in Equation \ref{eq:gaussian_kernel}, controls the smoothness of DNAMite's shape functions.
Larger values of $\phi$ yield smoother shape functions. 
For interactions, the final embedding is the weighted sum of the embeddings in a two-dimensional neighborhood of the input bin index pair.
For further details, see \citep{van2024dnamite}.

The DNAMite architecture has two key advantages over other NAM architectures.
First, discretizing all features allows for seamless handling of categorical features as well as missing values without any preprocessing.
As we demonstrate in Section \ref{usage}, separating the effect of missing values for each feature can lead to more detailed and representative model interpretations.
Second, many NAM architectures produce overly smooth shape functions, 
while the smoothness of DNAMite's shape function can be controlled by adjusting $\phi$.
As a result, DNAMite's shape functions often represent true shape functions more accurately than other NAMs \citep{van2024dnamite}.

\subsection{Feature Selection}
\label{sec:feature_selection}

\begin{figure}
    \centering
    \includegraphics[width=0.5\linewidth]{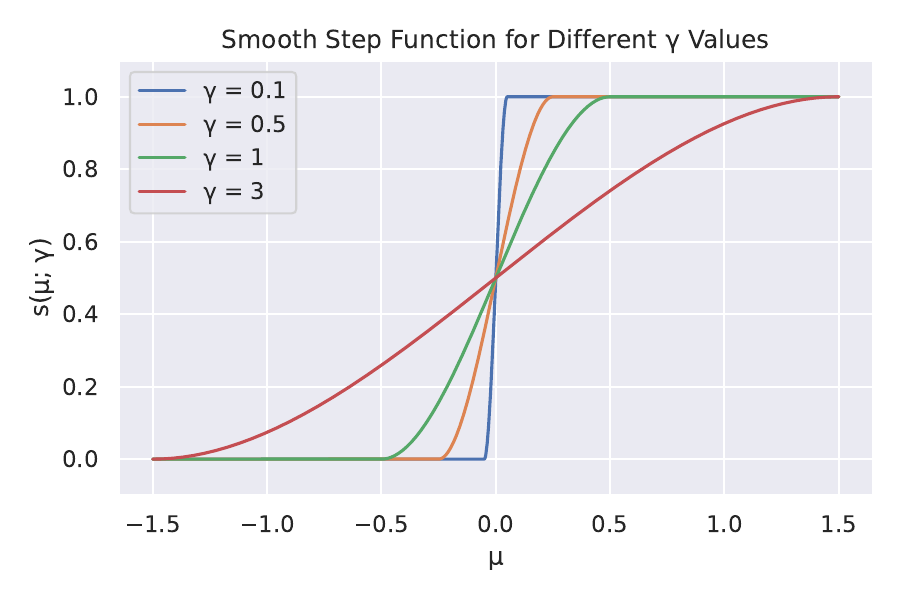}
    \caption{Visualization of the smooth-step function (Equation \ref{eq:smooth_step}) for different $\gamma$'s.}
    \label{fig:smooth_step}
\end{figure}

High multicollinearity (correlation between features) confounds the interpretation of linear model coefficients \citep{alin2010multicollinearity, kim2019multicollinearity, lavery2019number}. 
Similarly, strong multicollinearity 
can reduce the reliability of shape functions and their interpretation in additive models.
Several previous works have proposed using feature selection to address this problem \citep{van2024interpretable, ibrahim2024grand, yang2021gami, chang2021node, xu2023sparse, liu2020sparse}.

\textbf{dnamite} uses the method first proposed in \citep{ibrahim2024grand} for feature selection.
Starting from Equation \ref{eq:model_with_zs}, 
each feature gate $s(\mu_j), s(\mu_{j, k})$ contains a learnable parameter $\mu \in \mathbb{R}$ passed through the smooth-step function \citep{hazimeh2020tree}:
\begin{equation}
\label{eq:smooth_step}
s(\mu; \gamma) = 
    \begin{cases}
        0 & \text{ if } \mu \leq -\gamma/2 \\
        -\frac{2}{\gamma^{3}}\mu^3 + \frac{3}{2\gamma}\mu + \frac{1}{2} & \text{ if } -\gamma/2 \leq \mu \leq \gamma/2 \\
        1 & \text{ if } \mu \geq \gamma/2
    \end{cases},
\end{equation}
see Figure \ref{fig:smooth_step}.
The hyperparameter $\gamma$ controls how quickly the feature gates converge to $0$ and $1$. 
\textbf{dnamite} uses a regularizer $L_{\text{SS}}$ to encourage sparsity:
\begin{equation}
\label{eq:ss_loss}
    L_{\text{SS}} = \lambda \sum_j s(\mu_j).
\end{equation}
The hyperparameter $\lambda$ controls the level of feature sparsity in the resulting model, similar to the regularization parameter in a lasso model.
Any \textbf{dnamite} model can perform feature selection by adding 
$L_{\text{SS}}$ to the objective.
If a \textbf{dnamite} model includes interactions, then a similar regularizer can be used during interaction training to encourage sparsity in the interactions.

\subsection{Survival Analysis}

\textbf{dnamite} uses the approach outlined in \citep{van2024dnamite} to build NAMs for survival analysis.
First, $K$ evaluation times $t_1, t_2, \ldots, t_K$ are chosen using quantiles from the observed event times.
\textbf{dnamite} outputs a $K$-dimensional prediction $f(X) \in \mathbb{R}^K$ that estimates the conditional CDF at these evaluation times:
\begin{equation}
    \hat{P}(T \leq t_i \mid X) = f(X)_i, \quad i = 1, 2, \ldots, K.
\end{equation}
An element-wise sigmoid link function is used to ensure that all outputs represents valid probabilities.
The model is then trained using the Inverse Probability of Censoring Weighting (IPCW) loss
\begin{equation}
    \label{eq:ipcw_loss}
    \sum_{i=1}^n \sum_{k=1}^K \frac{\mathbbm{1}_{Z_i > t_k} \hat{p}(t_k)^2}{\hat{P}(C > t_k \mid X_i)} + \frac{\mathbbm{1}_{Z_i \leq t_k, \delta_i = 1} (1 - \hat{p}(t_k))^2}{\hat{P}(C > Z_i \mid X_i)}.
\end{equation}
This loss function is a proper scoring function for survival analysis under the assumption that $C \indep T \mid X$ \citep{rindt2022survival}. 
The censoring distribution $P(C > t \mid X)$ can be estimated using a Cox proportional hazards model fit before training the main model.
Under the additional assumption that $C \indep X$, this Cox model can be replaced with a Kaplan-Meier estimate for $P(C > t)$.

\section{Design Principles}
\label{sec:design}

\begin{figure}
    \centering
    \includegraphics[width=\linewidth]{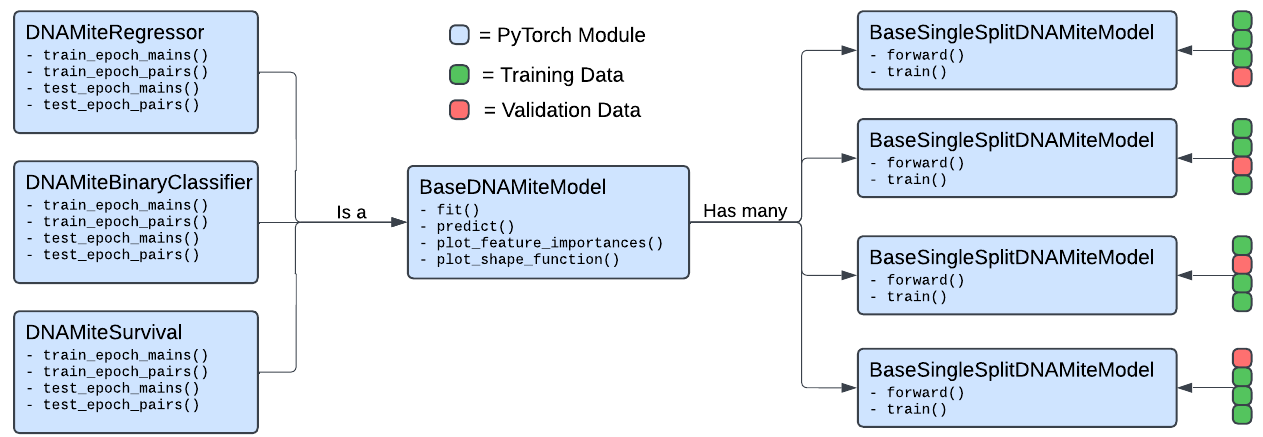}
    \caption{Class structure for \textbf{dnamite} models.}
    \label{fig:hierarchy}
\end{figure}

\textbf{dnamite} provides users with an easy-to-use interface for training and interacting with NAMs.
\textbf{dnamite} is designed in the style of scikit-learn, allowing users to train, extract information, and make predictions from a model with simple function calls directly from pandas dataframes.
This design contrasts with many neural network implementations that expect users to prepare specific data loaders and write custom training loops.
Although \textbf{dnamite} is written in PyTorch (thus scaling seamlessly on multi-core CPU or GPU hardware), users of \textbf{dnamite} do not need to write any PyTorch code.

Figure \ref{fig:hierarchy} shows the class structure used in \textbf{dnamite}. 
The base class \texttt{BaseDNAMiteModel} serves as the parent class for all \textbf{dnamite} models, implementing common functionality such as high-level functions for model fitting, predicting, and making explanation plots.
For each of the three supervised learning tasks that \textbf{dnamite} supports (regression, binary classification, survival analysis), \textbf{dnamite} implements a child class of \texttt{BaseDNAMiteModel} with task-specific functionality such as training loops.
When a \texttt{BaseDNAMiteModel} is trained, $k$ instances of \texttt{BaseSingleSplitDNAMiteModel} are trained, each on a different train/validation split. 
Each \texttt{BaseSingleSplitDNAMiteModel} implements the DNAMite architecture from Figure \ref{fig:architecture}.
Predictions, feature importances, and shape functions for a \texttt{BaseDNAMiteModel} are all computed by averaging the results across the single split models. 
This cross-validation approach allows \textbf{dnamite} to compute errors for feature importance and shape function plots, which is critical for assessing the reliability of these plots.

\hypertarget{usage}{%
\section{Usage}\label{usage}}

In this section, we demonstrate how to use \textbf{dnamite}
through an empirical example. 
Given the importance of interpretability and feature selection in healthcare applications, we focus our experiments on the
MIMIC III clinical database \citep{johnson2016mimic}, an electronic health records database from the Beth Israel
Deaconess Medical Center. 
Since MIMIC III is open
source and anonymized, it is a natural choice for reproducible
healthcare data science experiments.
For more usage information, see \textbf{dnamite}'s official documentation\footnote{\url{https://dnamite.readthedocs.io/}} and source code\footnote{\url{https://github.com/udellgroup/dnamite}}.

\hypertarget{data-preparation}{%
\subsection{Data Preparation}\label{data-preparation}}

\begin{table}[]
    \centering
    \scalebox{0.9}{
    \begin{tabular}{|llll|}
        \toprule
        Feature & Type & Missing Rate & Example Values \\
        \midrule
       Age & Continuous & 0.000 & 74.0, 23.0, 20.0  \\
       Gender & Categorical & 0.000 & `F', `F', `M' \\
       Ethnicity & Categorical & 0.000 & `WHITE', `BLACK',
        `UNKNOWN' \\
       Diastolic BP (mmHg) & Continuous & 0.173 & 55.107, 65.483,
        56.043 \\
       Inspired oxygen (\%) & Continuous & 0.814 & 52.0, 50.0, 70.0 \\
       GCS Total & Continuous & 0.522 & 11.889, 7.143, 14.526 \\
       GCS Eye Opening & Continuous & 0.175 & 3.667, 2.143, 3.842 \\
       GCS Motor Response & Continuous & 0.175 & 5.889, 3.143, 6.000 \\
       GCS Verbal Response & Continuous & 0.175 & 2.333, 1.857, 4.684 \\
       Glucose (mg/dL) & Continuous & 0.177 & 208.111, 154.611, 108.5 \\
       Heart Rate (bpm) & Continuous & 0.173 & 94.677, 135.775, 69.875 \\
       Height (in) & Continuous & 0.847 & 72.0, 60.0, 64.0 \\
       Mean Blood Pressure (mmHg) & Continuous & 0.173 & 72.071, 95.246, 78.159 \\
       Oxygen Saturation (\%) & Continuous & 0.172 & 97.906, 91.637, 99.25 \\
       Respiratory Rate (bpm) & Continuous & 0.173 & 13.565, 21.226, 17.125 \\
       Systolic Blood Pressure (mmHg) & Continuous & 0.173 & 110.262, 120.846, 122.391 \\
       Temperature (F) & Continuous & 0.186 & 98.7, 99.51, 98.12 \\
       Weight (kg) & Continuous & 0.700 & 51.2, 87.0, 78.3 \\
       pH & Continuous & 0.404 & 7.282, 7.279, 7.46\\
       Capillary refill rate & Categorical & 0.774 & `Brisk'', `Brisk', `Delayed' \\
       \bottomrule
    \end{tabular}
    }
    \caption{Features generated from MIMIC III for the benchmark dataset, following \citep{Harutyunyan2019}.}
    \label{tab:benchmark_features}
\end{table}

Preparing data for \textbf{dnamite} requires very little preprocessing.
All \textbf{dnamite} models ingest features as a pandas dataframe and labels as a pandas series or numpy array. 
Features can be categorical, continuous, or even missing, simplifying data processing: categorical features need not be encoded, continuous features need not be standardized,
and missing values need not be imputed as
\textbf{dnamite} naturally handles all of these data types.

To prepare the raw MIMIC III data for our usage demonstration, we
identify a cohort of 36,639 patients who are at least 18 years old and
have a minimum of 48 hours between their first admission and their last
discharge or death. 
Our focus is on mortality
prediction, with two subtasks: in-hospital mortality prediction as a
binary classification task, and time-to-death prediction from the index
time as a survival analysis problem.

We generate two mortality prediction datasets from the raw
MIMIC III data. 
The first is a comprehensive dataset with 757 features extracted from labevents and
chartevents. 
The second is a benchmark
dataset consisting of one feature for each of the 17 clinical variables
used in the MIMIC III benchmark \citep{Harutyunyan2019}, see Table \ref{tab:benchmark_features}.
Most features align with well-known medical concepts. One lesser-known concept is the Glasgow Coma Scale (GCS), which assesses consciousness through eye, motor, and verbal responses. Lower values in these categories indicate reduced responsiveness.

We can generate these datasets from the raw MIMIC III data using \textbf{dnamite}'s \texttt{fetch\_mimic} function.
    \begin{tcolorbox}[breakable, size=fbox, boxrule=1pt, pad at break*=1mm,colback=cellbackground, colframe=cellborder]
\begin{Verbatim}[commandchars=\\\{\}]
\PY{k+kn}{from} \PY{n+nn}{dnamite}\PY{n+nn}{.}\PY{n+nn}{datasets} \PY{k+kn}{import} \PY{n}{fetch\PYZus{}mimic}
\PY{n}{data\PYZus{}path} \PY{o}{=} \PY{l+s+s2}{\PYZdq{}}\PY{l+s+s2}{/home/mvanness/mimic/mimic\PYZus{}raw\PYZus{}data/}\PY{l+s+s2}{\PYZdq{}}
\PY{n}{cohort\PYZus{}data}\PY{p}{,} \PY{n}{benchmark\PYZus{}data} \PY{o}{=} \PY{n}{fetch\PYZus{}mimic}\PY{p}{(}
    \PY{n}{data\PYZus{}path}\PY{o}{=}\PY{n}{data\PYZus{}path}\PY{p}{,} 
    \PY{n}{return\PYZus{}benchmark\PYZus{}data}\PY{o}{=}\PY{k+kc}{True}
\PY{p}{)}
\end{Verbatim}
\end{tcolorbox}
See the package documentation for complete details on data generation.
The input file path should be
adjusted to point to a directory with the raw MIMIC III csv files, which may be obtained from physionet\footnote{\url{https://physionet.org/content/mimiciii/1.4/}}.

    \hypertarget{basic-usage}{%
\subsection{Basic Usage}\label{basic-usage}}

We start by importing standard packages and defining a few useful variables.

    \begin{tcolorbox}[breakable, size=fbox, boxrule=1pt, pad at break*=1mm,colback=cellbackground, colframe=cellborder]
\begin{Verbatim}[commandchars=\\\{\}]
\PY{k+kn}{import} \PY{n+nn}{numpy} \PY{k}{as} \PY{n+nn}{np}
\PY{k+kn}{import} \PY{n+nn}{pandas} \PY{k}{as} \PY{n+nn}{pd}  
\PY{k+kn}{import} \PY{n+nn}{matplotlib}\PY{n+nn}{.}\PY{n+nn}{pyplot} \PY{k}{as} \PY{n+nn}{plt}
\PY{k+kn}{import} \PY{n+nn}{seaborn} \PY{k}{as} \PY{n+nn}{sns}
\PY{n}{sns}\PY{o}{.}\PY{n}{set\PYZus{}theme}\PY{p}{(}\PY{p}{)}
\PY{k+kn}{from} \PY{n+nn}{sklearn}\PY{n+nn}{.}\PY{n+nn}{model\PYZus{}selection} \PY{k+kn}{import} \PY{n}{train\PYZus{}test\PYZus{}split}
\PY{k+kn}{import} \PY{n+nn}{torch} 
\PY{k+kn}{import} \PY{n+nn}{os}
\PY{n}{device} \PY{o}{=} \PY{n}{torch}\PY{o}{.}\PY{n}{device}\PY{p}{(}\PY{l+s+s1}{\PYZsq{}}\PY{l+s+s1}{cuda:0}\PY{l+s+s1}{\PYZsq{}} \PY{k}{if} \PY{n}{torch}\PY{o}{.}\PY{n}{cuda}\PY{o}{.}\PY{n}{is\PYZus{}available}\PY{p}{(}\PY{p}{)} \PY{k}{else} \PY{l+s+s1}{\PYZsq{}}\PY{l+s+s1}{cpu}\PY{l+s+s1}{\PYZsq{}}\PY{p}{)}
\PY{n}{seed} \PY{o}{=} \PY{l+m+mi}{10}
\end{Verbatim}
\end{tcolorbox}

The most basic use of \textbf{dnamite} is as an additive model for regression or classification. This paper demonstrates binary classification, but regression is very similar. 
The following code prepares the benchmark in-hospital mortality binary classification dataset.

    \begin{tcolorbox}[breakable, size=fbox, boxrule=1pt, pad at break*=1mm,colback=cellbackground, colframe=cellborder]
\begin{Verbatim}[commandchars=\\\{\}]
\PY{n}{X\PYZus{}benchmark} \PY{o}{=} \PY{n}{benchmark\PYZus{}data}\PY{o}{.}\PY{n}{drop}\PY{p}{(}
    \PY{n}{columns}\PY{o}{=}\PY{p}{[}\PY{l+s+s2}{\PYZdq{}}\PY{l+s+s2}{SUBJECT\PYZus{}ID}\PY{l+s+s2}{\PYZdq{}}\PY{p}{,} \PY{l+s+s2}{\PYZdq{}}\PY{l+s+s2}{HOSPITAL\PYZus{}EXPIRE\PYZus{}FLAG}\PY{l+s+s2}{\PYZdq{}}\PY{p}{]}
\PY{p}{)}
\PY{n}{y\PYZus{}benchmark} \PY{o}{=} \PY{n}{benchmark\PYZus{}data}\PY{p}{[}\PY{l+s+s2}{\PYZdq{}}\PY{l+s+s2}{HOSPITAL\PYZus{}EXPIRE\PYZus{}FLAG}\PY{l+s+s2}{\PYZdq{}}\PY{p}{]}
\PY{n}{X\PYZus{}train\PYZus{}benchmark}\PY{p}{,} \PY{n}{X\PYZus{}test\PYZus{}benchmark}\PY{p}{,} \PYZbs{}
\PY{n}{y\PYZus{}train\PYZus{}benchmark}\PY{p}{,} \PY{n}{y\PYZus{}test\PYZus{}benchmark} \PY{o}{=} \PY{n}{train\PYZus{}test\PYZus{}split}\PY{p}{(}
    \PY{n}{X\PYZus{}benchmark}\PY{p}{,}
    \PY{n}{y\PYZus{}benchmark}\PY{p}{,}
    \PY{n}{test\PYZus{}size}\PY{o}{=}\PY{l+m+mf}{0.2}\PY{p}{,}
    \PY{n}{random\PYZus{}state}\PY{o}{=}\PY{n}{seed}\PY{p}{,}
\PY{p}{)}
\end{Verbatim}
\end{tcolorbox}

Fitting a binary classification \textbf{dnamite} model follows the standard
scikit-learn style API, using the \texttt{DNAMiteBinaryClassifier} class. 
After initialization, we train a \textbf{dnamite} model using the \texttt{fit} function.

    \begin{tcolorbox}[breakable, size=fbox, boxrule=1pt, pad at break*=1mm,colback=cellbackground, colframe=cellborder]
\begin{Verbatim}[commandchars=\\\{\}]
\PY{k+kn}{from} \PY{n+nn}{dnamite}\PY{n+nn}{.}\PY{n+nn}{models} \PY{k+kn}{import} \PY{n}{DNAMiteBinaryClassifier}
\PY{n}{model} \PY{o}{=} \PY{n}{DNAMiteBinaryClassifier}\PY{p}{(}
   \PY{n}{n\PYZus{}features}\PY{o}{=}\PY{n}{X\PYZus{}benchmark}\PY{o}{.}\PY{n}{shape}\PY{p}{[}\PY{l+m+mi}{1}\PY{p}{]}\PY{p}{,} 
   \PY{n}{random\PYZus{}state}\PY{o}{=}\PY{n}{seed}\PY{p}{,}
   \PY{n}{device}\PY{o}{=}\PY{n}{device}\PY{p}{,}
   \PY{n}{num\PYZus{}pairs}\PY{o}{=}\PY{l+m+mi}{5}
\PY{p}{)}
\PY{n}{model}\PY{o}{.}\PY{n}{fit}\PY{p}{(}\PY{n}{X\PYZus{}train\PYZus{}benchmark}\PY{p}{,} \PY{n}{y\PYZus{}train\PYZus{}benchmark}\PY{p}{)}
\end{Verbatim}
\end{tcolorbox}

\texttt{DNAMiteBinaryClassifier} has several parameters that can be set at
initialization, the full list of which is available in the package documentation. The only required parameter is
\texttt{n\_features}, which should be set to the number of features in the dataset. Three optional parameters are also specified:
\begin{itemize}
    \item \texttt{random\_state}: a random seed for controlling reproducibility.
    \item \texttt{device}: a PyTorch device or string used to initialize a PyTorch
    device. The default is ``cpu''. Passing a cuda device or string name
    triggers GPU training, which is more efficient.
    \item  \texttt{num\_pairs}:
    the number of pairwise interactions to use.
    To select exactly $k$ interactions, \textbf{dnamite} keeps the $k$ interactions with the largest gates $s(\mu_{j, k})$. The default is zero, which excludes interactions
    from the model.
\end{itemize}

When we call the \texttt{fit} function, several steps happen under the
hood. 
\begin{enumerate}
    \item \textbf{Identifying Data Types}: \textbf{dnamite} first automatically determines the data type of each
    feature in the input training data. The three possible data types are
    continuous, binary, and categorical. 
    \item \textbf{Defining Bins}: \textbf{dnamite} defines feature bins for
    each feature in order to prepare for discretization and embedding. 
    Two optional parameters can be passed to a \textbf{dnamite}
    model that affect the binning of continuous features. First,
    \texttt{max\_bins} controls the maximum number of bins a feature can
    have. The default is 32. Second,
    \texttt{min\_samples\_per\_bin} sets a lower bound on the number of
    samples from the training dataset that must belong to each bin. The
    default is the minimum of 50 and 1\% of the training data samples. Since
    each bin is assigned an embedding, this safeguard ensures that each
    embedding has enough data to properly train.
    \item \textbf{Fitting Single Split Models}: After defining data types and feature bins, \textbf{dnamite} starts fitting
    single split DNAMite models, each to a different train/validation split (see Figure \ref{fig:hierarchy}).
    The optional parameter \texttt{n\_val\_splits}, which defaults to 5,
    controls how many data splits are used. For a given split, \textbf{dnamite} uses
    the defined bins to transform the training and validation data so that
    all values are replaced by their corresponding bin index. This
    transformed data is fed to PyTorch data loaders to prepare the data for
    training. If \texttt{num\_pairs} is set to a value greater than 0, then
    pairs are selected using the first split and remain the same for all
    splits.
\end{enumerate}

We can save and load \texttt{dnamite} models using the standard PyTorch saving and loading functions.

    \begin{tcolorbox}[breakable, size=fbox, boxrule=1pt, pad at break*=1mm,colback=cellbackground, colframe=cellborder]
\begin{Verbatim}[commandchars=\\\{\}]
torch.save(model, \PY{l+s+s2}{\PYZdq{}}\PY{l+s+s2}{mimic\PYZus{}benchmark\PYZus{}model.pth}\PY{l+s+s2}{\PYZdq{}}\PY{p})
\PY{n}{model} \PY{o}{=} \PY{n}{torch}\PY{o}{.}\PY{n}{load}\PY{p}{(}\PY{l+s+s2}{\PYZdq{}}\PY{l+s+s2}{mimic\PYZus{}benchmark\PYZus{}model.pth}\PY{l+s+s2}{\PYZdq{}}\PY{p}{)}
\end{Verbatim}
\end{tcolorbox}

Next, we use the 
\texttt{predict\_proba} function to get probabilistic predictions from
the model, which are needed to compute the AUC score.

    \begin{tcolorbox}[breakable, size=fbox, boxrule=1pt, pad at break*=1mm,colback=cellbackground, colframe=cellborder]
\begin{Verbatim}[commandchars=\\\{\}]
\PY{k+kn}{from} \PY{n+nn}{sklearn}\PY{n+nn}{.}\PY{n+nn}{metrics} \PY{k+kn}{import} \PY{n}{roc\PYZus{}auc\PYZus{}score}
\PY{n}{preds} \PY{o}{=} \PY{n}{model}\PY{o}{.}\PY{n}{predict\PYZus{}proba}\PY{p}{(}\PY{n}{X\PYZus{}test\PYZus{}benchmark}\PY{p}{)}
\PY{n}{dnamite\PYZus{}auc} \PY{o}{=} \PY{n}{np}\PY{o}{.}\PY{n}{round}\PY{p}{(}\PY{n}{roc\PYZus{}auc\PYZus{}score}\PY{p}{(}\PY{n}{y\PYZus{}test\PYZus{}benchmark}\PY{p}{,} \PY{n}{preds}\PY{p}{)}\PY{p}{,} \PY{l+m+mi}{4}\PY{p}{)}
\PY{n}{dnamite\PYZus{}auc}
\end{Verbatim}
\end{tcolorbox}

\begin{Verbatim}[commandchars=\\\{\}]
0.8521
\end{Verbatim}

After training a \textbf{dnamite} model, 
we can visualize
feature importances and shape
functions with very little
additional computation. 
We first use the \texttt{plot\_feature\_importances}
function to plot the feature importance scores of the
top features in descending order. 

    \begin{tcolorbox}[breakable, size=fbox, boxrule=1pt, pad at break*=1mm,colback=cellbackground, colframe=cellborder]
\begin{Verbatim}[commandchars=\\\{\}]
\PY{n}{model}\PY{o}{.}\PY{n}{plot\PYZus{}feature\PYZus{}importances}\PY{p}{(}{)}
\end{Verbatim}
\end{tcolorbox}

\begin{center}
    \includegraphics[width=0.5\linewidth]{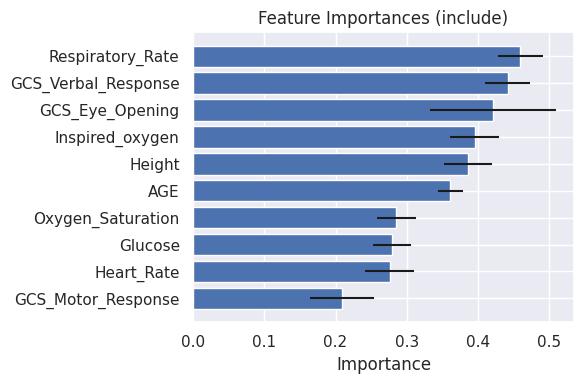}
\end{center}




Each bar represents the mean
importance across validation splits, with error bars representing
confidence intervals. The optional parameter \texttt{n\_features} can be
used to change how many features are plotted. If the model uses pairwise
interactions, these are also each assigned a score and eligible for inclusion in the plot.
For example, if feature ``A''
and feature ``B'' had an important interaction, it would be displayed as ``A
\textbar\textbar{} B''.

A second optional parameter \texttt{missing\_bin} changes how the
missing bin is used in the feature importance calculation.
The following are the options for setting this parameter.
\begin{itemize}
    \item \texttt{"include"}: the default, which treats the
    missing bin the same as all other bins. This option is acceptable for
    most use cases but can result in misleadingly high importances if a
    feature's missing bin has a high absolute score due to informative
    missingness. 
    \item \texttt{"ignore"}: disregard the missing bin,
    i.e. average the absolute value of the shape function only for samples
    where the feature is not missing. This option is useful if a user does
    not want missing values to influence feature importances, or if
    diagnosing and analyzing informative missingness is not important.
    \item \texttt{"stratify"}: compute separate feature importance scores for
    missing and observed. The observed score is the same as the score when
    using ``ignore'', and the missing score is the absolute value of the missing
    bin's output. The feature order is determined by the observed score.
    This setting allows the user to separate the importance of a
    feature into missing and observed components, similar to computing
    feature importances for a model after using the missing indicator method
    \citep{van2023missing}.
\end{itemize}

Below we demonstrate the non-default options for our trained \textbf{dnamite} model.

    \begin{tcolorbox}[breakable, size=fbox, boxrule=1pt, pad at break*=1mm,colback=cellbackground, colframe=cellborder]
\begin{Verbatim}[commandchars=\\\{\}]
\PY{n}{model}\PY{o}{.}\PY{n}{plot\PYZus{}feature\PYZus{}importances}\PY{p}{(}\PY{n}{missing\PYZus{}bin}\PY{o}{=}\PY{l+s+s2}{\PYZdq{}}\PY{l+s+s2}{ignore}\PY{l+s+s2}{\PYZdq{}}\PY{p}{)}
\PY{n}{model}\PY{o}{.}\PY{n}{plot\PYZus{}feature\PYZus{}importances}\PY{p}{(}\PY{n}{missing\PYZus{}bin}\PY{o}{=}\PY{l+s+s2}{\PYZdq{}}\PY{l+s+s2}{stratify}\PY{l+s+s2}{\PYZdq{}}\PY{p}{)}
\end{Verbatim}
\end{tcolorbox}

\begin{center}
    \begin{minipage}[c]{0.4\textwidth}
        \centering
        \includegraphics[width=\textwidth]{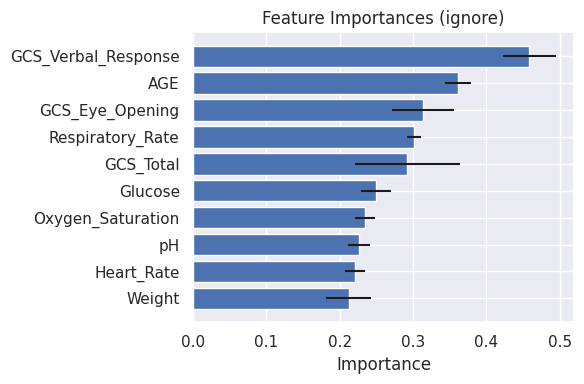}
    \end{minipage}
    \begin{minipage}[c]{0.58\textwidth}
        \centering
        \includegraphics[width=\textwidth]{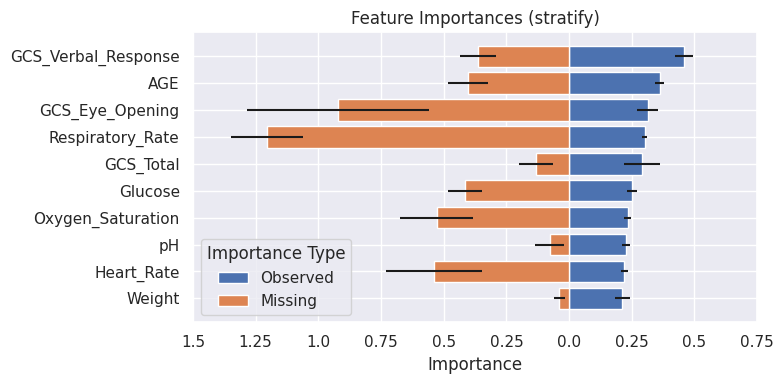}
    \end{minipage}
\end{center}

Respiratory rate is the most important feature using the default
combined score, but is only the fourth most important feature by
observed score. 
The stratified importance plot sheds light on this difference: respiratory rate has a very large missing score, thereby increasing its overall importance score.
This example illustrates how the stratified importance plot can help users better understand the role of each feature.
Importantly,
such stratification is possible due to DNAMite's embedding module, and cannot be achieved with other NAM architectures.

Next, we use the
\texttt{plot\_shape\_function} function to plot estimated shape functions for
selected features.
    \begin{tcolorbox}[breakable, size=fbox, boxrule=1pt, pad at break*=1mm,colback=cellbackground, colframe=cellborder]
\begin{Verbatim}[commandchars=\\\{\}]
\PY{n}{model}\PY{o}{.}\PY{n}{plot\PYZus{}shape\PYZus{}function}\PY{p}{(}\PY{p}{[}
    \PY{l+s+s2}{\PYZdq{}}\PY{l+s+s2}{AGE}\PY{l+s+s2}{\PYZdq{}}\PY{p}{,}
    \PY{l+s+s2}{\PYZdq{}}\PY{l+s+s2}{GCS\PYZus{}Verbal\PYZus{}Response}\PY{l+s+s2}{\PYZdq{}}\PY{p}{,}
    \PY{l+s+s2}{\PYZdq{}}\PY{l+s+s2}{Respiratory\PYZus{}Rate}\PY{l+s+s2}{\PYZdq{}}\PY{p}{,}
\PY{p}{]}\PY{p}{,} \PY{n}{plot\PYZus{}missing\PYZus{}bin}\PY{o}{=}\PY{k+kc}{True}\PY{p}{)}
\end{Verbatim}
\end{tcolorbox}

\begin{center}
\includegraphics[width=\linewidth]{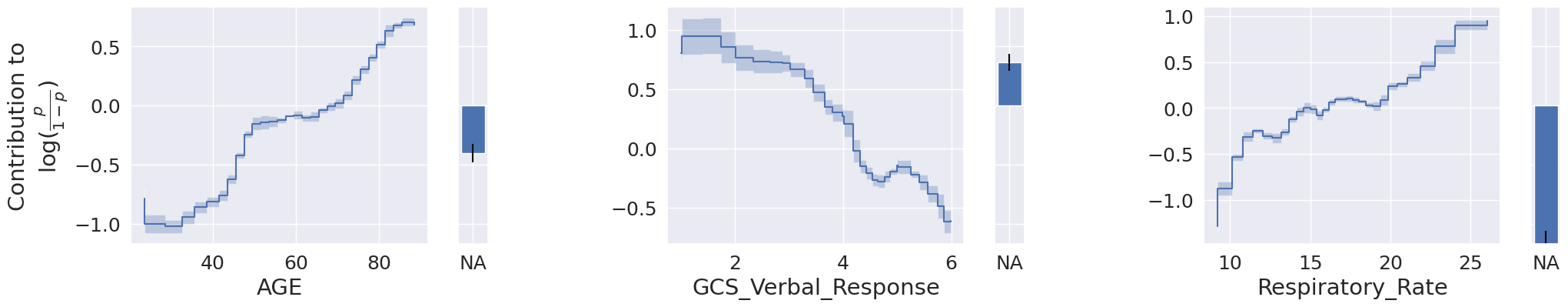}
\end{center}

Similar to the feature importance
plot, the error region in these shape functions represents pointwise
confidence intervals. The function's first parameter accepts either a
single string or a list of strings representing the name(s) of the
feature(s) to plot. A second optional parameter
\texttt{plot\_missing\_bin} defaults to \texttt{False} but can be set to
\texttt{True} to visualize the missing bin score.
    
Age and respiratory rate, both critical indicators of a patient's health, show positive correlations with in-hospital mortality.
Conversely, GCS verbal response exhibits a negative correlation with in-hospital mortality, indicating that better verbal responses are associated with lower mortality rates. 
Additionally, patients with a missing respiratory rate measurement are at much lower risk of in-hospital mortality. This missing bin score exemplifies informative missingness, as patients who are critically ill are more likely to have their respiratory rate monitored.

We can also visualize a model's feature interactions.
We first list the interactions used by the model, which are stored in 
\texttt{model.selected\_pairs}.

    \begin{tcolorbox}[breakable, size=fbox, boxrule=1pt, pad at break*=1mm,colback=cellbackground, colframe=cellborder]
\begin{Verbatim}[commandchars=\\\{\}]
\PY{n}{model}\PY{o}{.}\PY{n}{selected\PYZus{}pairs}
\end{Verbatim}
\end{tcolorbox}

            \begin{tcolorbox}[breakable, size=fbox, boxrule=.5pt, pad at break*=1mm, opacityfill=0]
\begin{Verbatim}[commandchars=\\\{\}]
[['GCS\_Eye\_Opening', 'Systolic\_Blood\_Pressure'],
 ['Oxygen\_Saturation', 'pH'],
 ['Systolic\_Blood\_Pressure', 'pH'],
 ['AGE', 'Temperature'],
 ['Heart\_Rate', 'pH']]
\end{Verbatim}
\end{tcolorbox}

We can plot one of these interactions using the \texttt{plot\_pair\_shape\_function} function.

\begin{tcolorbox}[breakable, size=fbox, boxrule=1pt, pad at break*=1mm,colback=cellbackground, colframe=cellborder]
\begin{Verbatim}[commandchars=\\\{\}]
\PY{n}{model}\PY{o}{.}\PY{n}{plot\PYZus{}pair\PYZus{}shape\PYZus{}function}\PY{p}{(}
    \PY{l+s+s2}{\PYZdq{}}\PY{l+s+s2}{GCS\PYZus{}Eye\PYZus{}Opening}\PY{l+s+s2}{\PYZdq{}}\PY{p}{,} 
    \PY{l+s+s2}{\PYZdq{}}\PY{l+s+s2}{Systolic\PYZus{}Blood\PYZus{}Pressure}\PY{l+s+s2}{\PYZdq{}}
\PY{p}{)}
\end{Verbatim}
\end{tcolorbox}

\begin{center}
\includegraphics[width=0.5\linewidth]{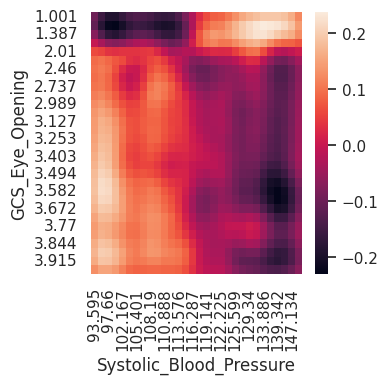}
\end{center}


This interaction plot between eye opening and systolic blood pressure (SBP) shows two regions with increased risk: very low eye opening with mid-to-high SBP as well as low SBP and mid-to-high eye opening.
However, this interaction has relatively low importance: it was not a top 10 feature in any of our feature importance plots.
Interestingly, we have found that \textbf{dnamite} models without any interactions often still have excellent predictive performance.
    
    \hypertarget{comparison-to-existing-packages}{%
\subsubsection{Comparison to Existing
Packages}\label{comparison-to-existing-packages}}

\begin{figure}
\centering
\includegraphics[width=\linewidth]{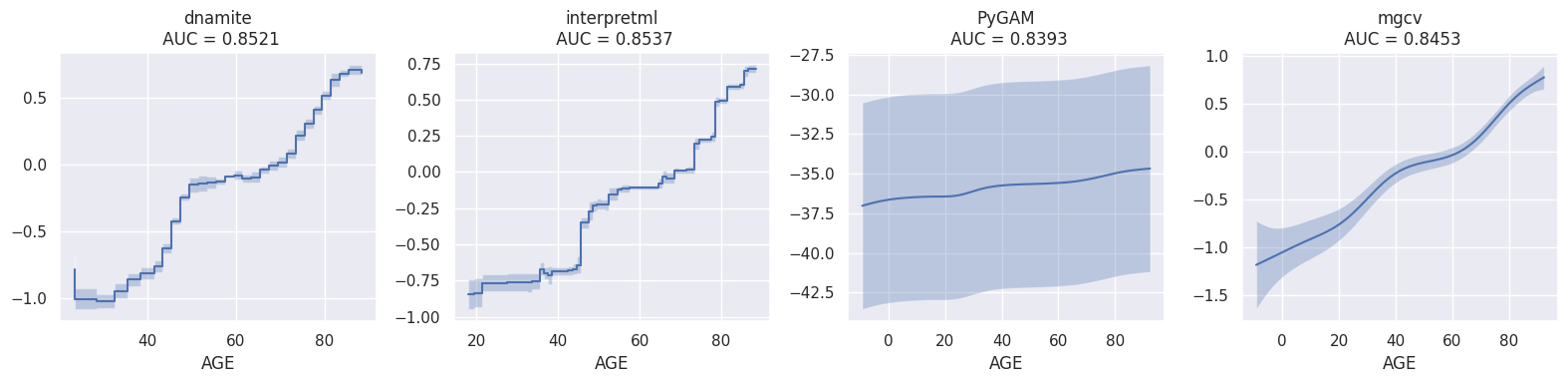}
\caption{Comparison of \textbf{dnamite} to other additive model packages. For each package, the shape function for age is given as well as the test AUC.}
\label{fig:dnamite_comparison}
\end{figure}

Figure \ref{fig:dnamite_comparison} compares the shape function for age among \textbf{dnamite} and three other packages from Table \ref{tab:related_work}: interpretml, PyGAM, and mgcv.  
Both \textbf{dnamite} and interpretml produce similar shape functions with comparable test AUC values.  
mgcv's shape function is similar in overall shape but is slightly smoother and has a slightly worse test AUC.  
Overall, these three packages provide similar value for standard regression/classification modeling.
In contrast, PyGAM generates a shape function that is completely uninterpretable despite having a test AUC that is only marginally worse than the others.

    \hypertarget{feature-selection}{%
\subsection{Feature Selection}\label{feature-selection}}

When feature selection is desired, \textbf{dnamite} allows users to
select a subset of features before training a model. 
\textbf{dnamite} performs feature selection as a separate initial step, ensuring consistent feature use across all single split models.

We demonstrate \textbf{dnamite}'s ability to do feature selection using the complete cohort dataset.
    \begin{tcolorbox}[breakable, size=fbox, boxrule=1pt, pad at break*=1mm,colback=cellbackground, colframe=cellborder]
\begin{Verbatim}[commandchars=\\\{\}]
\PY{n}{cohort\PYZus{}data} \PY{o}{=} \PY{n}{cohort\PYZus{}data}\PY{o}{.}\PY{n}{drop}\PY{p}{(}\PY{p}{[}
   \PY{l+s+s2}{\PYZdq{}}\PY{l+s+s2}{SUBJECT\PYZus{}ID}\PY{l+s+s2}{\PYZdq{}}\PY{p}{,}
   \PY{l+s+s2}{\PYZdq{}}\PY{l+s+s2}{DOB}\PY{l+s+s2}{\PYZdq{}}\PY{p}{,}
   \PY{l+s+s2}{\PYZdq{}}\PY{l+s+s2}{DOD}\PY{l+s+s2}{\PYZdq{}}\PY{p}{,}
   \PY{l+s+s2}{\PYZdq{}}\PY{l+s+s2}{ADMITTIME}\PY{l+s+s2}{\PYZdq{}}\PY{p}{,}
   \PY{l+s+s2}{\PYZdq{}}\PY{l+s+s2}{DEATH\PYZus{}TIME}\PY{l+s+s2}{\PYZdq{}}\PY{p}{,}
   \PY{l+s+s2}{\PYZdq{}}\PY{l+s+s2}{CENSOR\PYZus{}TIME}\PY{l+s+s2}{\PYZdq{}}\PY{p}{,}
   \PY{l+s+s2}{\PYZdq{}}\PY{l+s+s2}{LAST\PYZus{}DISCHTIME}\PY{l+s+s2}{\PYZdq{}}\PY{p}{,}
   \PY{l+s+s2}{\PYZdq{}}\PY{l+s+s2}{index\PYZus{}time}\PY{l+s+s2}{\PYZdq{}}
\PY{p}{]}\PY{p}{,} \PY{n}{axis}\PY{o}{=}\PY{l+m+mi}{1}\PY{p}{)}
\PY{n}{X\PYZus{}cls} \PY{o}{=} \PY{n}{cohort\PYZus{}data}\PY{o}{.}\PY{n}{drop}\PY{p}{(}\PY{p}{[}\PY{l+s+s2}{\PYZdq{}}\PY{l+s+s2}{HOSPITAL\PYZus{}EXPIRE\PYZus{}FLAG}\PY{l+s+s2}{\PYZdq{}}\PY{p}{,} \PY{l+s+s2}{\PYZdq{}}\PY{l+s+s2}{event}\PY{l+s+s2}{\PYZdq{}}\PY{p}{,} \PY{l+s+s2}{\PYZdq{}}\PY{l+s+s2}{time}\PY{l+s+s2}{\PYZdq{}}\PY{p}{]}\PY{p}{,} \PY{n}{axis}\PY{o}{=}\PY{l+m+mi}{1}\PY{p}{)}
\PY{n}{y\PYZus{}cls} \PY{o}{=} \PY{n}{cohort\PYZus{}data}\PY{p}{[}\PY{l+s+s2}{\PYZdq{}}\PY{l+s+s2}{HOSPITAL\PYZus{}EXPIRE\PYZus{}FLAG}\PY{l+s+s2}{\PYZdq{}}\PY{p}{]}
\PY{n}{X\PYZus{}train\PYZus{}cls}\PY{p}{,} \PY{n}{X\PYZus{}test\PYZus{}cls}\PY{p}{,} \PY{n}{y\PYZus{}train\PYZus{}cls}\PY{p}{,} \PY{n}{y\PYZus{}test\PYZus{}cls} \PY{o}{=} \PY{n}{train\PYZus{}test\PYZus{}split}\PY{p}{(}
   \PY{n}{X\PYZus{}cls}\PY{p}{,} \PY{n}{y\PYZus{}cls}\PY{p}{,} \PY{n}{test\PYZus{}size}\PY{o}{=}\PY{l+m+mf}{0.2}\PY{p}{,} \PY{n}{random\PYZus{}state}\PY{o}{=}\PY{n}{seed}
\PY{p}{)}
\end{Verbatim}
\end{tcolorbox}
Since this dataset contains 757 features, feature selection is required to obtain a low-dimensional additive model.
After initializing a \texttt{DNAMiteBinaryClassifier} model as before,
we use the \texttt{select\_features} function to select features and interactions.

    \begin{tcolorbox}[breakable, size=fbox, boxrule=1pt, pad at break*=1mm,colback=cellbackground, colframe=cellborder]
\begin{Verbatim}[commandchars=\\\{\}]
\PY{k+kn}{from} \PY{n+nn}{dnamite}\PY{n+nn}{.}\PY{n+nn}{models} \PY{k+kn}{import} \PY{n}{DNAMiteBinaryClassifier}
\PY{n}{model} \PY{o}{=} \PY{n}{DNAMiteBinaryClassifier}\PY{p}{(}
   \PY{n}{random\PYZus{}state}\PY{o}{=}\PY{n}{seed}\PY{p}{,}
   \PY{n}{n\PYZus{}features}\PY{o}{=}\PY{n}{X\PYZus{}cls}\PY{o}{.}\PY{n}{shape}\PY{p}{[}\PY{l+m+mi}{1}\PY{p}{]}\PY{p}{,} 
   \PY{n}{device}\PY{o}{=}\PY{n}{device}\PY{p}{,}
\PY{p}{)}
\PY{n}{model}\PY{o}{.}\PY{n}{select\PYZus{}features}\PY{p}{(}
   \PY{n}{X\PYZus{}train\PYZus{}cls}\PY{p}{,} 
   \PY{n}{y\PYZus{}train\PYZus{}cls}\PY{p}{,} 
   \PY{n}{reg\PYZus{}param}\PY{o}{=}\PY{l+m+mf}{0.2}\PY{p}{,}
   \PY{n}{select\PYZus{}pairs}\PY{o}{=}\PY{k+kc}{True}\PY{p}{,}
   \PY{n}{pair\PYZus{}reg\PYZus{}param}\PY{o}{=}\PY{l+m+mf}{0.0015}\PY{p}{,}
\PY{p}{)}
\end{Verbatim}
\end{tcolorbox}

Similar to the \texttt{fit}
function, the \texttt{select\_features} function accepts training
features and labels as its first two arguments. A third required
parameter \texttt{reg\_param} controls the strength of the feature
sparsity regularizer, i.e.~\(\lambda\) in Equation \ref{eq:ss_loss}. 
Larger values of $\lambda$ will result in fewer selected features. Additionally,
\texttt{select\_features} accepts the following optional parameters that
are used to further control the feature selection process. 
\begin{itemize}
    \item \texttt{select\_pairs}: whether or not to select a list of interactions
    along with the selected features. Defaults to \texttt{False}.
    \item \texttt{pair\_reg\_param}: similar to \texttt{reg\_param} but
    controlling the strength of pair selection.
    \item \texttt{gamma}: controls the steepness of the smooth-step function as
    described in Equation \ref{eq:smooth_step} and Figure \ref{fig:smooth_step}. Smaller values reduce the number of
    iterations required before feature pruning starts. The default value,
    which we have found works well in many settings, is \begin{equation}
    \texttt{gamma} = \min\left( \frac{N}{B} \cdot \frac{1}{250} \cdot \frac{16}{d}, 1 \right),
    \end{equation} where \(N\) is the number of training samples, \(B\) is
    the batch size, and \(d\) is the hidden dimension. The intuition is that
    as \(N / B\) (number of iterations per epoch) increases, the model will usually run more iterations before convergence. Hence \texttt{gamma} should also be increased so that features are not pruned too early.
    \item \texttt{pair\_gamma}: similar to
    \texttt{gamma} but for pairs selection. Defaults to
    \texttt{gamma\ /\ 4}.
\end{itemize}

    After feature selection, the selected features and pairs are stored in
\texttt{model.selected\_feats} and \texttt{model.selected\_pairs}
respectively.

    \begin{tcolorbox}[breakable, size=fbox, boxrule=1pt, pad at break*=1mm,colback=cellbackground, colframe=cellborder]
\begin{Verbatim}[commandchars=\\\{\}]
\PY{n+nb}{print}\PY{p}{(}\PY{l+s+s2}{\PYZdq{}}\PY{l+s+s2}{Number of selected features:}\PY{l+s+s2}{\PYZdq{}}\PY{p}{,} \PY{n+nb}{len}\PY{p}{(}\PY{n}{model}\PY{o}{.}\PY{n}{selected\PYZus{}feats}\PY{p}{)}\PY{p}{)}
\PY{n+nb}{print}\PY{p}{(}\PY{n}{model}\PY{o}{.}\PY{n}{selected\PYZus{}feats}\PY{p}{)}
\end{Verbatim}
\end{tcolorbox}

    \begin{Verbatim}[commandchars=\\\{\}]
Number of selected features: 12
['AGE',
 'lab\_Lactate\_mmol/L',
 'lab\_Anion\_Gap\_mEq/L',
 'lab\_Bicarbonate\_mEq/L',
 'lab\_Sodium\_mEq/L',
 'lab\_RDW\_\%',
 'chart\_Level\_of\_Conscious',
 'chart\_Activity',
 'chart\_Code\_Status',
 'chart\_Gag\_Reflex',
 'chart\_Orientation',
 'chart\_Spontaneous\_Movement']
    \end{Verbatim}

    \begin{tcolorbox}[breakable, size=fbox, boxrule=1pt, pad at break*=1mm,colback=cellbackground, colframe=cellborder]
\begin{Verbatim}[commandchars=\\\{\}]
\PY{n+nb}{print}\PY{p}{(}\PY{l+s+s2}{\PYZdq{}}\PY{l+s+s2}{Number of selected pairs:}\PY{l+s+s2}{\PYZdq{}}\PY{p}{,} \PY{n+nb}{len}\PY{p}{(}\PY{n}{model}\PY{o}{.}\PY{n}{selected\PYZus{}pairs}\PY{p}{)}\PY{p}{)}
\PY{n+nb}{print}\PY{p}{(}\PY{n}{model}\PY{o}{.}\PY{n}{selected\PYZus{}pairs}\PY{p}{)}
\end{Verbatim}
\end{tcolorbox}

    \begin{Verbatim}[commandchars=\\\{\}]
Number of selected pairs: 1
[['lab\_Lactate\_mmol/L', 'lab\_Bicarbonate\_mEq/L']]
    \end{Verbatim}

Now, we call the \texttt{fit} function as before,
and the model will automatically use the selected features and interactions.

    \begin{tcolorbox}[breakable, size=fbox, boxrule=1pt, pad at break*=1mm,colback=cellbackground, colframe=cellborder]
\begin{Verbatim}[commandchars=\\\{\}]
\PY{n}{model}\PY{o}{.}\PY{n}{fit}\PY{p}{(}\PY{n}{X\PYZus{}train\PYZus{}cls}\PY{p}{,} \PY{n}{y\PYZus{}train\PYZus{}cls}\PY{p}{)}
\end{Verbatim}
\end{tcolorbox}

We obtain model predictions similarly, with the model again
automatically using the selected features and interactions.

    \begin{tcolorbox}[breakable, size=fbox, boxrule=1pt, pad at break*=1mm,colback=cellbackground, colframe=cellborder]
\begin{Verbatim}[commandchars=\\\{\}]
\PY{k+kn}{from} \PY{n+nn}{sklearn}\PY{n+nn}{.}\PY{n+nn}{metrics} \PY{k+kn}{import} \PY{n}{roc\PYZus{}auc\PYZus{}score}
\PY{n}{preds} \PY{o}{=} \PY{n}{model}\PY{o}{.}\PY{n}{predict\PYZus{}proba}\PY{p}{(}\PY{n}{X\PYZus{}test\PYZus{}cls}\PY{p}{)}
\PY{n}{selection\PYZus{}auc} \PY{o}{=} \PY{n}{np}\PY{o}{.}\PY{n}{round}\PY{p}{(}\PY{n}{roc\PYZus{}auc\PYZus{}score}\PY{p}{(}\PY{n}{y\PYZus{}test\PYZus{}cls}\PY{p}{,} \PY{n}{preds}\PY{p}{)}\PY{p}{,} \PY{l+m+mi}{4}\PY{p}{)}
\PY{n+nb}{print}\PY{p}{(}\PY{l+s+sa}{f}\PY{l+s+s2}{\PYZdq{}}\PY{l+s+s2}{AUC with automated feature selection: }\PY{l+s+si}{\PYZob{}}\PY{n}{selection\PYZus{}auc}\PY{l+s+si}{\PYZcb{}}\PY{l+s+s2}{\PYZdq{}}\PY{p}{)}
\PY{n+nb}{print}\PY{p}{(}\PY{l+s+sa}{f}\PY{l+s+s2}{\PYZdq{}}\PY{l+s+s2}{AUC with manual feature selection: }\PY{l+s+si}{\PYZob{}}\PY{n}{dnamite\PYZus{}auc}\PY{l+s+si}{\PYZcb{}}\PY{l+s+s2}{\PYZdq{}}\PY{p}{)}
\end{Verbatim}
\end{tcolorbox}

\begin{Verbatim}[commandchars=\\\{\}]
AUC with automated feature selection: 0.8792
AUC with manual feature selection: 0.8521
\end{Verbatim}
        
Compared to the features selected in the benchmark, the AUC of this model is
approximately 0.025 higher, demonstrating the potential lift in predictive performance due to automated feature selection.
We see the benefits of combining
feature selection and additive modeling within the same package and model structure,
a combination that distinguishes \textbf{dnamite} from previous packages for
additive modeling or feature selection.

As before, we can visualize feature importances and shape functions after training.

    \begin{tcolorbox}[breakable, size=fbox, boxrule=1pt, pad at break*=1mm,colback=cellbackground, colframe=cellborder]
\begin{Verbatim}[commandchars=\\\{\}]
\PY{n}{model}\PY{o}{.}\PY{n}{plot\PYZus{}feature\PYZus{}importances}\PY{p}{(}\PY{n}{missing\PYZus{}bin}\PY{o}{=}\PY{l+s+s2}{\PYZdq{}}\PY{l+s+s2}{stratify}\PY{l+s+s2}{\PYZdq{}}\PY{p}{)}
\end{Verbatim}
\end{tcolorbox}

\begin{center}
\includegraphics[width=0.7\linewidth]{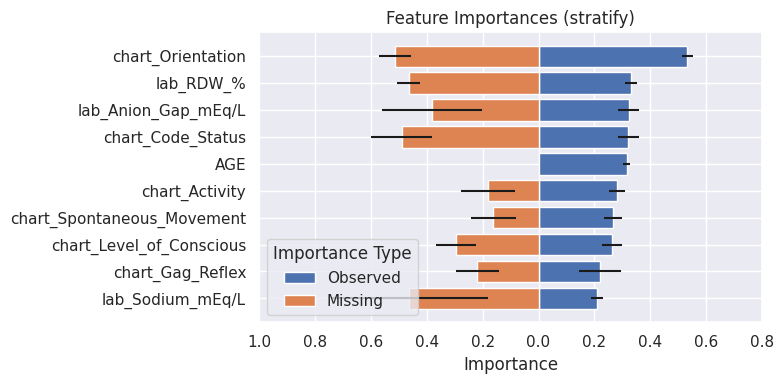}
\end{center}

\begin{tcolorbox}[breakable, size=fbox, boxrule=1pt, pad at break*=1mm,colback=cellbackground, colframe=cellborder]
\begin{Verbatim}[commandchars=\\\{\}]
\PY{n}{model}\PY{o}{.}\PY{n}{plot\PYZus{}shape\PYZus{}function}\PY{p}{(}\PY{p}{[}
    \PY{l+s+s2}{\PYZdq{}}\PY{l+s+s2}{chart\PYZus{}Level\PYZus{}of\PYZus{}Conscious}\PY{l+s+s2}{\PYZdq{}}\PY{p}{,}
    \PY{l+s+s2}{\PYZdq{}}\PY{l+s+s2}{chart\PYZus{}Orientation}\PY{l+s+s2}{\PYZdq{}}\PY{p}{,}
    \PY{l+s+s2}{\PYZdq{}}\PY{l+s+s2}{AGE}\PY{l+s+s2}{\PYZdq{}}\PY{p}{,}
\PY{p}{]}\PY{p}{)}
\end{Verbatim}
\end{tcolorbox}

\begin{center}
\includegraphics[width=\linewidth]{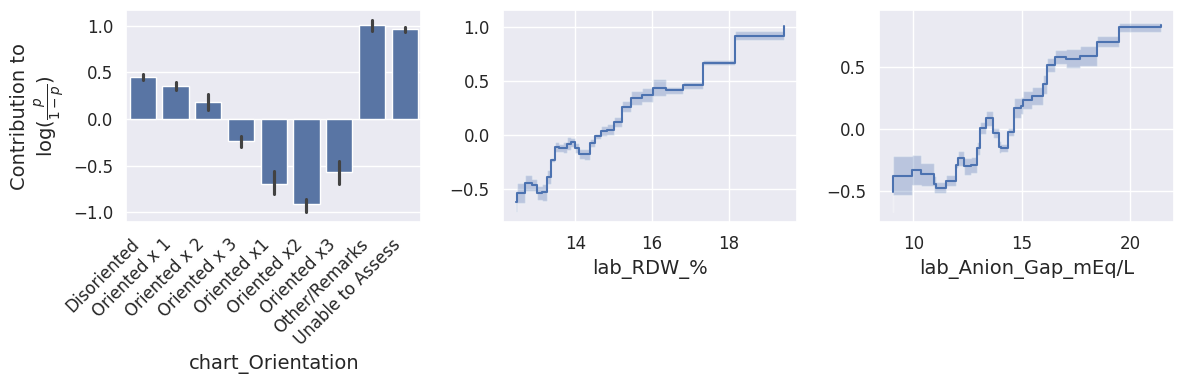}
\end{center}
    
Interestingly, all of the top features except age are different in this
model compared to the benchmark model. Many of these new features
represent direct or indirect measures of overall health and/or
seriousness of care. 
For example, the most important feature is orientation, which assesses a patient's cognitive awareness.
Patients who are more disoriented, need special remarks, or cannot have their orientation assessed are (unsurprisingly) at higher mortality risk.
Without algorithmic feature selection, MIMIC users might never realize that any feature has such strong correlation with in-hospital mortality.
Moreover, if a user wishes to remove a feature like orientation, 
believing that it may leak information about the label, they can easily fit a new \textbf{dnamite} model that excludes that feature. 
Users can also choose to combine manual and algorithmic feature selection to refine their model.

    \hypertarget{comparison-to-existing-packages}{%
\subsubsection{Comparison to Existing
Packages}\label{comparison-to-existing-packages}}

As illustrated in Table \ref{tab:related_work}, the only other additive model package that supports feature selection is PiML. 
However, PiML's feature selection capabilities are notably inferior to those of \textbf{dnamite}. 
Firstly, GAMI-Net \citep{yang2021gami} is the only model in PiML that implements feature selection. 
GAMI-Net initially fits all features and subsequently selects the top $k$ features based on shape function variation. 
This strategy is inefficient because the model must fully estimate shape functions for all features before performing any selection.
In contrast, a \textbf{dnamite} model can prune a feature as soon as the feature's gate is reduced to zero. 
Additionally, GAMI-Net's approach to feature selection is more heuristic compared to \textbf{dnamite}'s, as \textbf{dnamite} integrates feature selection directly into the optimization process. Furthermore, \textbf{dnamite}'s feature selection has been demonstrated to outperform GAMI-Net \citep{ibrahim2024grand}. 
PiML also lacks comprehensive user guides or examples to illustrate how feature selection can be executed using GAMI-Net; the PiML API includes a function, \texttt{fine\_tune\_selected}, for pruning features but does not provide examples of its application.

    \hypertarget{survival-analysis}{%
\subsection{Survival Analysis}\label{survival-analysis}}

Training and analyzing a \textbf{dnamite} model for survival analysis is
similar to the workflow for regression or classification.
Following scikit-survival \citep{polsterl2020scikit},
\textbf{dnamite} expects labels prepared as
a structured NumPy array with dtype
\texttt{{[}(\textquotesingle{}event\textquotesingle{},\ \textquotesingle{}bool\textquotesingle{}),\ (\textquotesingle{}time\textquotesingle{},\ \textquotesingle{}float\textquotesingle{}){]}}.
The \texttt{event} field should store \texttt{True} when the event of interest is
observed and \texttt{False} when the sample is censored. The \texttt{time} field
should store the time-to-event for observed samples and
time-to-censoring for censored samples.
For the MIMIC mortality data, patients are censored with their last discharge time if their death time is not recorded in the dataset.

    \begin{tcolorbox}[breakable, size=fbox, boxrule=1pt, pad at break*=1mm,colback=cellbackground, colframe=cellborder]
\begin{Verbatim}[commandchars=\\\{\}]
\PY{n}{X\PYZus{}surv} \PY{o}{=} \PY{n}{cohort\PYZus{}data}\PY{o}{.}\PY{n}{drop}\PY{p}{(}
   \PY{p}{[}\PY{l+s+s2}{\PYZdq{}}\PY{l+s+s2}{HOSPITAL\PYZus{}EXPIRE\PYZus{}FLAG}\PY{l+s+s2}{\PYZdq{}}\PY{p}{,} \PY{l+s+s2}{\PYZdq{}}\PY{l+s+s2}{event}\PY{l+s+s2}{\PYZdq{}}\PY{p}{,} \PY{l+s+s2}{\PYZdq{}}\PY{l+s+s2}{time}\PY{l+s+s2}{\PYZdq{}}\PY{p}{]}\PY{p}{,} 
   \PY{n}{axis}\PY{o}{=}\PY{l+m+mi}{1}
\PY{p}{)}
\PY{n}{y\PYZus{}surv} \PY{o}{=} \PY{n}{np}\PY{o}{.}\PY{n}{empty}\PY{p}{(}
   \PY{n}{dtype}\PY{o}{=}\PY{p}{[}\PY{p}{(}\PY{l+s+s1}{\PYZsq{}}\PY{l+s+s1}{event}\PY{l+s+s1}{\PYZsq{}}\PY{p}{,} \PY{l+s+s1}{\PYZsq{}}\PY{l+s+s1}{bool}\PY{l+s+s1}{\PYZsq{}}\PY{p}{)}\PY{p}{,} \PY{p}{(}\PY{l+s+s1}{\PYZsq{}}\PY{l+s+s1}{time}\PY{l+s+s1}{\PYZsq{}}\PY{p}{,} \PY{l+s+s1}{\PYZsq{}}\PY{l+s+s1}{float64}\PY{l+s+s1}{\PYZsq{}}\PY{p}{)}\PY{p}{]}\PY{p}{,} 
   \PY{n}{shape}\PY{o}{=}\PY{p}{(}\PY{n}{cohort\PYZus{}data}\PY{o}{.}\PY{n}{shape}\PY{p}{[}\PY{l+m+mi}{0}\PY{p}{]}\PY{p}{,}\PY{p}{)}
\PY{p}{)}
\PY{n}{y\PYZus{}surv}\PY{p}{[}\PY{l+s+s2}{\PYZdq{}}\PY{l+s+s2}{event}\PY{l+s+s2}{\PYZdq{}}\PY{p}{]} \PY{o}{=} \PY{n}{cohort\PYZus{}data}\PY{p}{[}\PY{l+s+s2}{\PYZdq{}}\PY{l+s+s2}{event}\PY{l+s+s2}{\PYZdq{}}\PY{p}{]}\PY{o}{.}\PY{n}{values}
\PY{n}{y\PYZus{}surv}\PY{p}{[}\PY{l+s+s2}{\PYZdq{}}\PY{l+s+s2}{time}\PY{l+s+s2}{\PYZdq{}}\PY{p}{]} \PY{o}{=} \PY{n}{np}\PY{o}{.}\PY{n}{clip}\PY{p}{(}\PY{n}{cohort\PYZus{}data}\PY{p}{[}\PY{l+s+s2}{\PYZdq{}}\PY{l+s+s2}{time}\PY{l+s+s2}{\PYZdq{}}\PY{p}{]}\PY{o}{.}\PY{n}{values}\PY{p}{,} \PY{l+m+mf}{1e\PYZhy{}5}\PY{p}{,} \PY{n}{np}\PY{o}{.}\PY{n}{inf}\PY{p}{)}
\PY{n}{X\PYZus{}train\PYZus{}surv}\PY{p}{,} \PY{n}{X\PYZus{}test\PYZus{}surv}\PY{p}{,} \PY{n}{y\PYZus{}train\PYZus{}surv}\PY{p}{,} \PY{n}{y\PYZus{}test\PYZus{}surv} \PY{o}{=} \PY{n}{train\PYZus{}test\PYZus{}split}\PY{p}{(}
   \PY{n}{X\PYZus{}surv}\PY{p}{,} \PY{n}{y\PYZus{}surv}\PY{p}{,} \PY{n}{test\PYZus{}size}\PY{o}{=}\PY{l+m+mf}{0.2}\PY{p}{,} \PY{n}{random\PYZus{}state}\PY{o}{=}\PY{n}{seed}
\PY{p}{)}
\end{Verbatim}
\end{tcolorbox}

\textbf{dnamite} uses the class \texttt{DNAMiteSurvival} for survival analysis
models. 
We can fit a feature-sparse \textbf{dnamite} survival model using the same functions as before.

    \begin{tcolorbox}[breakable, size=fbox, boxrule=1pt, pad at break*=1mm,colback=cellbackground, colframe=cellborder]
\begin{Verbatim}[commandchars=\\\{\}]
\PY{k+kn}{from} \PY{n+nn}{dnamite}\PY{n+nn}{.}\PY{n+nn}{models} \PY{k+kn}{import} \PY{n}{DNAMiteSurvival}
\PY{n}{model} \PY{o}{=} \PY{n}{DNAMiteSurvival}\PY{p}{(}
    \PY{n}{random\PYZus{}state}\PY{o}{=}\PY{n}{seed}\PY{p}{,}
    \PY{n}{n\PYZus{}features}\PY{o}{=}\PY{n}{X\PYZus{}surv}\PY{o}{.}\PY{n}{shape}\PY{p}{[}\PY{l+m+mi}{1}\PY{p}{]}\PY{p}{,}
    \PY{n}{device}\PY{o}{=}\PY{n}{device}\PY{p}{,} 
\PY{p}{)}
\PY{n}{model}\PY{o}{.}\PY{n}{select\PYZus{}features}\PY{p}{(}
    \PY{n}{X\PYZus{}train\PYZus{}surv}\PY{p}{,} 
    \PY{n}{y\PYZus{}train\PYZus{}surv}\PY{p}{,} 
    \PY{n}{select\PYZus{}pairs}\PY{o}{=}\PY{k+kc}{True}\PY{p}{,}
    \PY{n}{reg\PYZus{}param}\PY{o}{=}\PY{l+m+mf}{0.75}\PY{p}{,}
    \PY{n}{pair\PYZus{}reg\PYZus{}param}\PY{o}{=}\PY{l+m+mf}{0.005}\PY{p}{,}
\PY{p}{)}
\PY{n+nb}{print}\PY{p}{(}\PY{l+s+s2}{\PYZdq{}}\PY{l+s+s2}{Number of selected features:}\PY{l+s+s2}{\PYZdq{}}\PY{p}{,} \PY{n+nb}{len}\PY{p}{(}\PY{n}{model}\PY{o}{.}\PY{n}{selected\PYZus{}feats}\PY{p}{)}\PY{p}{)}
\PY{n+nb}{print}\PY{p}{(}\PY{l+s+s2}{\PYZdq{}}\PY{l+s+s2}{Number of selected pairs:}\PY{l+s+s2}{\PYZdq{}}\PY{p}{,} \PY{n+nb}{len}\PY{p}{(}\PY{n}{model}\PY{o}{.}\PY{n}{selected\PYZus{}pairs}\PY{p}{)}\PY{p}{)}
\PY{n}{model}\PY{o}{.}\PY{n}{fit}\PY{p}{(}\PY{n}{X\PYZus{}train\PYZus{}surv}\PY{p}{,} \PY{n}{y\PYZus{}train\PYZus{}surv}\PY{p}{)}
\end{Verbatim}
\end{tcolorbox}

    \begin{Verbatim}[commandchars=\\\{\}]
Number of selected features: 17
Number of selected pairs: 4
    \end{Verbatim}

       \textbf{dnamite} survival models optimize the IPCW loss given in Equation \ref{eq:ipcw_loss}.
By default, the censoring distribution is estimated using a Kaplan-Meier
model. To use a Cox model instead, the optional parameter
\texttt{censor\_estimator} should be set to \texttt{"cox"} at initialization.

Unlike previous classification models, a \textbf{dnamite} survival model yields
interpretations for specific evaluation times. 
In the
\texttt{plot\_feature\_importances}
function, we add a parameter \texttt{eval\_times} to specify either a
single evaluation time or a list of evaluation times for which feature
importance scores should be plotted.

    \begin{tcolorbox}[breakable, size=fbox, boxrule=1pt, pad at break*=1mm,colback=cellbackground, colframe=cellborder]
\begin{Verbatim}[commandchars=\\\{\}]
\PY{n}{model}\PY{o}{.}\PY{n}{plot\PYZus{}feature\PYZus{}importances}\PY{p}{(}\PY{n}{eval\PYZus{}times}\PY{o}{=}\PY{p}{[}\PY{l+m+mi}{7}\PY{p}{,} \PY{l+m+mi}{365}\PY{p}{]}\PY{p}{,} \PY{n}{missing\PYZus{}bin}\PY{o}{=}\PY{l+s+s2}{\PYZdq{}}\PY{l+s+s2}{stratify}\PY{l+s+s2}{\PYZdq{}}\PY{p}{)}
\end{Verbatim}
\end{tcolorbox}

\begin{center}
\includegraphics[width=\linewidth]{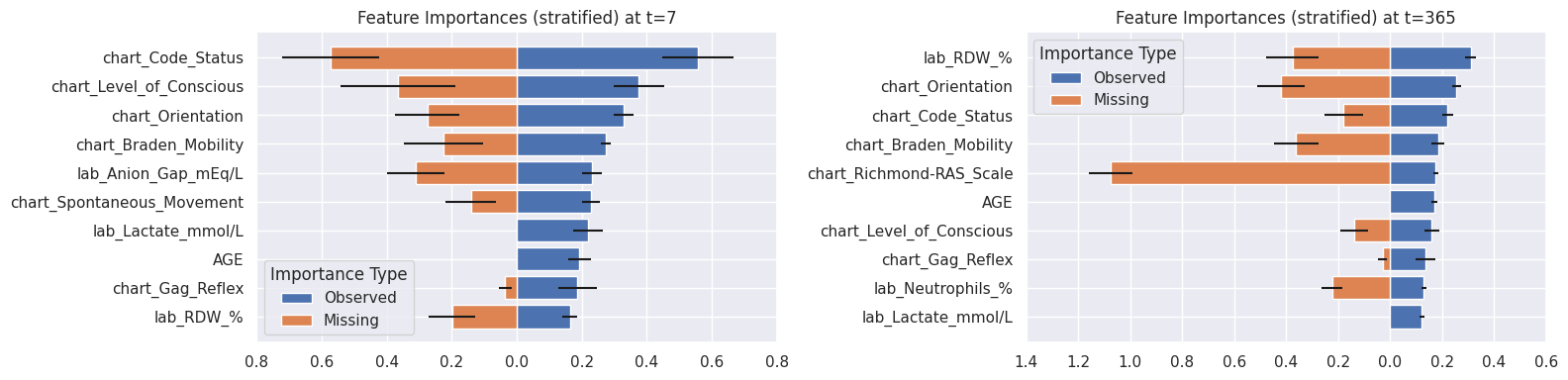}
\end{center}

These importances illustrate interesting differences in the risk factors for
short-term (next 7 days) and long-term (next year) mortality risk. For
short-term mortality, the most important feature is code status, which describes how much medical intervention should be taken in the event of a medical emergency.
Another important feature relative is Anion Gap, with elevated levels being a well-documented predictor of imminent mortality risk \citep{brenner1985clinical}. 
Conversely, Red Cell Distribution Width (RDW) is relatively more significant for predicting long-term mortality risk \citep{wang2011red}.

Similarly, we call the \texttt{plot_shape_function} function with parameter \texttt{eval_times} to plot shape functions for features at specific evaluation times.
    
    \begin{tcolorbox}[breakable, size=fbox, boxrule=1pt, pad at break*=1mm,colback=cellbackground, colframe=cellborder]
\begin{Verbatim}[commandchars=\\\{\}]
\PY{n}{model}\PY{o}{.}\PY{n}{plot\PYZus{}shape\PYZus{}function}\PY{p}{(}
    \PY{p}{[}\PY{l+s+s2}{\PYZdq{}}\PY{l+s+s2}{chart\PYZus{}Code\PYZus{}Status}\PY{l+s+s2}{\PYZdq{}}\PY{p}{,} \PY{l+s+s2}{\PYZdq{}}\PY{l+s+s2}{chart\PYZus{}Level\PYZus{}of\PYZus{}Conscious}\PY{l+s+s2}{\PYZdq{}}\PY{p}{,} \PY{l+s+s2}{\PYZdq{}}\PY{l+s+s2}{lab\PYZus{}RDW\PYZus{}}\PY{l+s+s2}{\PYZpc{}}\PY{l+s+s2}{\PYZdq{}}\PY{p}{]}\PY{p}{,}
    \PY{n}{eval\PYZus{}times}\PY{o}{=}\PY{p}{[}\PY{l+m+mi}{7}\PY{p}{,} \PY{l+m+mi}{365}\PY{p}{]}
\PY{p}{)}
\end{Verbatim}
\end{tcolorbox}

\begin{center}
\includegraphics[width=\linewidth]{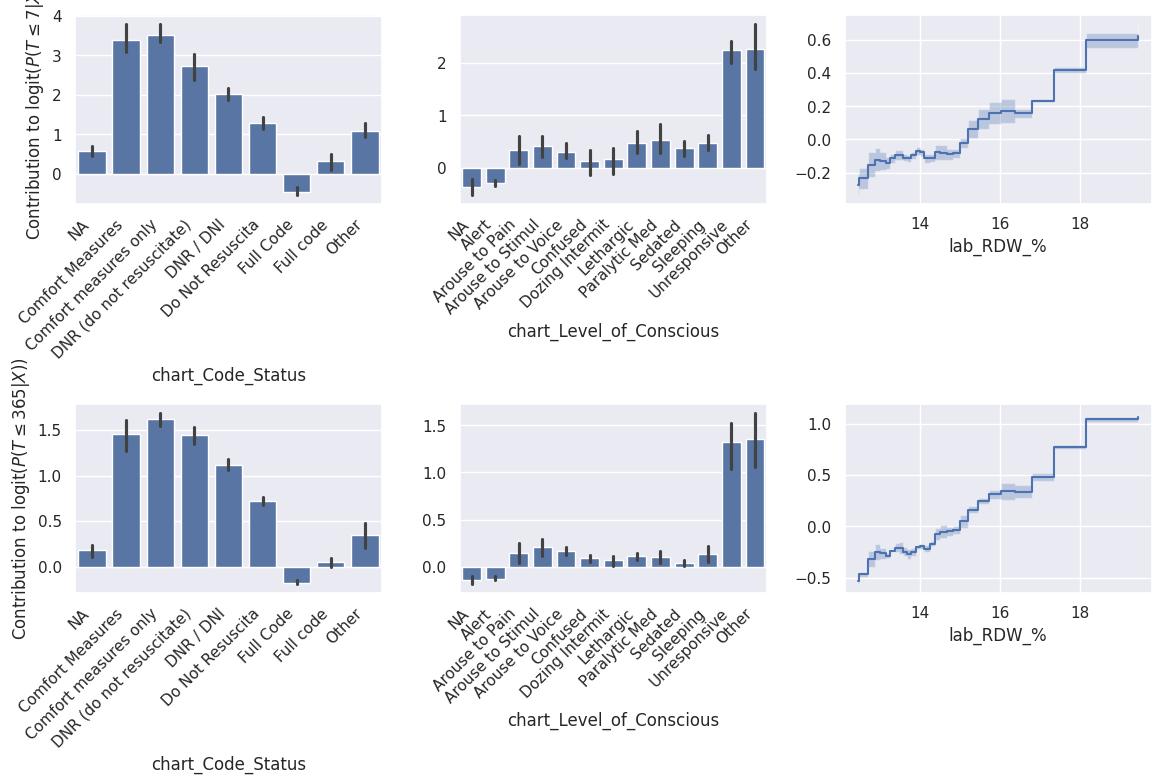}
\end{center}

The patterns of these shape functions remain consistent at different evaluation times, although their magnitudes change to reflect varying levels of importance across time.

Assessing the calibration of survival models is important as calibration ensures that survival predictions have valid probabilistic interpretations \citep{austin2020graphical}. 
\textbf{dnamite} supports calibration assessment via calibration plots. 
To make these calibration plots, \textbf{dnamite} first partitions
samples into bins using quantiles of the predicted CDF estimates at a given evaluation time. 
Within
each bin, a Kaplan-Meier estimator is used to estimate the true CDF
score, which is plotted against the bin's average predicted CDF from the
model. A trained \textbf{dnamite} survival model can call the function
\texttt{make\_calibration\_plot} to assess calibration. The
function accepts features, labels, and a list of evaluation times.

    \begin{tcolorbox}[breakable, size=fbox, boxrule=1pt, pad at break*=1mm,colback=cellbackground, colframe=cellborder]
\begin{Verbatim}[commandchars=\\\{\}]
\PY{n}{model}\PY{o}{.}\PY{n}{make\PYZus{}calibration\PYZus{}plot}\PY{p}{(}\PY{n}{X\PYZus{}test\PYZus{}surv}\PY{p}{,} \PY{n}{y\PYZus{}test\PYZus{}surv}\PY{p}{,} \PY{n}{eval\PYZus{}times}\PY{o}{=}\PY{p}{[}\PY{l+m+mi}{7}\PY{p}{,} \PY{l+m+mi}{365}\PY{p}{]}\PY{p}{)}
\end{Verbatim}
\end{tcolorbox}

\begin{center}
\includegraphics[width=0.8\linewidth]{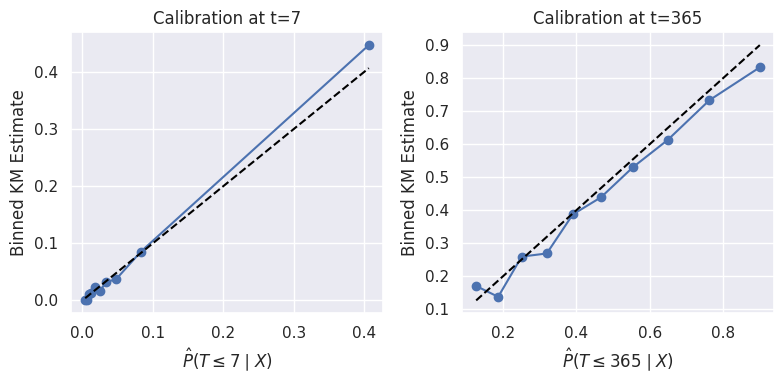}
\end{center}
    
These plots illustrate that the trained model is reasonably calibrated since the points are all close to the $y = x$ line.
If the model were to appear uncalibrated, 
a possible cause is that $C \notindep X$, so that the IPCW loss in Equation \ref{eq:ipcw_loss} is improper. 
In that case, we recommend retraining the
model setting \texttt{censor\_estimator} to \texttt{"cox"}
to see if relaxing the assumption to $C \indep T \mid X$ improves calibration.

    \hypertarget{comparison-to-existing-packages}{%
\subsubsection{Comparison to Existing
Packages}\label{comparison-to-existing-packages}}

As shown in Table \ref{tab:related_work}, there are no existing Python packages that
support training additive models for survival analysis. The R packages
gam and mgcv support survival analysis, but with several limitations
compared to \textbf{dnamite}. First, both packages only support training with the
Cox proportional hazards loss, constraining these models to the
proportional hazards assumption that \textbf{dnamite} avoids. Second, neither R
package supports feature selection, which is a core component of
\textbf{dnamite}. Third, neither package can handle missing values natively.
Instead, they would require missing value imputation, and so they cannot identify
the impact of missing values in feature importances and shape functions.

    \hypertarget{advanced-usage}{%
\subsection{Advanced Usage}\label{advanced-usage}}

    \hypertarget{controlling-smoothness}{%
\subsubsection{Controlling Smoothness}\label{controlling-smoothness}}

As discussed in Section \ref{sec:architecture}, the parameter $\phi$ from Equation \ref{eq:gaussian_kernel} 
controls the smoothness of \textbf{dnamite}'s shape functions.
Exposing $\phi$
as an optional parameter gives users more control over shape function smoothness. 
A user who wants simple shape functions should set $\phi$ to a large value.
Conversely, a user who wants more complicated shape functions at the risk of overfitting should set $\phi$ to a small value.

The optional parameter \texttt{kernel\_weight} sets $\phi$ in \textbf{dnamite} models at initialization.
Below we demonstrate the impact of changing
\texttt{kernel\_weight} on the \texttt{GCS\_Verbal\_Response}
feature in the benchmark binary classification dataset.

    \begin{tcolorbox}[breakable, size=fbox, boxrule=1pt, pad at break*=1mm,colback=cellbackground, colframe=cellborder]
\begin{Verbatim}[commandchars=\\\{\}]
\PY{n}{fig}\PY{p}{,} \PY{n}{axes} \PY{o}{=} \PY{n}{plt}\PY{o}{.}\PY{n}{subplots}\PY{p}{(}\PY{l+m+mi}{1}\PY{p}{,} \PY{l+m+mi}{5}\PY{p}{,} \PY{n}{figsize}\PY{o}{=}\PY{p}{(}\PY{l+m+mi}{20}\PY{p}{,} \PY{l+m+mi}{4}\PY{p}{)}\PY{p}{)}
\PY{n}{kernel\PYZus{}weights} \PY{o}{=} \PY{p}{[}\PY{l+m+mi}{0}\PY{p}{,} \PY{l+m+mi}{1}\PY{p}{,} \PY{l+m+mi}{5}\PY{p}{,} \PY{l+m+mi}{10}\PY{p}{,} \PY{l+m+mi}{100}\PY{p}{]}
\PY{k}{for} \PY{n}{kw}\PY{p}{,} \PY{n}{ax} \PY{o+ow}{in} \PY{n+nb}{zip}\PY{p}{(}\PY{n}{kernel\PYZus{}weights}\PY{p}{,} \PY{n}{axes}\PY{p}{)}\PY{p}{:}
    \PY{n}{model} \PY{o}{=} \PY{n}{DNAMiteBinaryClassifier}\PY{p}{(}
        \PY{n}{random\PYZus{}state}\PY{o}{=}\PY{n}{seed}\PY{p}{,}
        \PY{n}{n\PYZus{}features}\PY{o}{=}\PY{n}{X\PYZus{}benchmark}\PY{o}{.}\PY{n}{shape}\PY{p}{[}\PY{l+m+mi}{1}\PY{p}{]}\PY{p}{,} 
        \PY{n}{device}\PY{o}{=}\PY{n}{device}\PY{p}{,}
        \PY{n}{kernel\PYZus{}weight}\PY{o}{=}\PY{n}{kw}\PY{p}{,}
    \PY{p}{)}
    \PY{n}{model}\PY{o}{.}\PY{n}{fit}\PY{p}{(}\PY{n}{X\PYZus{}train\PYZus{}benchmark}\PY{p}{,} \PY{n}{y\PYZus{}train\PYZus{}benchmark}\PY{p}{)}
    \PY{n}{model}\PY{o}{.}\PY{n}{plot\PYZus{}shape\PYZus{}function}\PY{p}{(}\PY{l+s+s2}{\PYZdq{}}\PY{l+s+s2}{GCS\PYZus{}Verbal\PYZus{}Response}\PY{l+s+s2}{\PYZdq{}}\PY{p}{,} \PY{n}{axes}\PY{o}{=}\PY{p}{[}\PY{n}{ax}\PY{p}{]}\PY{p}{)}
    \PY{n}{ax}\PY{o}{.}\PY{n}{set\PYZus{}title}\PY{p}{(}\PY{l+s+sa}{f}\PY{l+s+s2}{\PYZdq{}}\PY{l+s+s2}{Kernel Weight = }\PY{l+s+si}{\PYZob{}}\PY{n}{kw}\PY{l+s+si}{\PYZcb{}}\PY{l+s+s2}{\PYZdq{}}\PY{p}{)}
\PY{n}{plt}\PY{o}{.}\PY{n}{tight\PYZus{}layout}\PY{p}{(}\PY{p}{)}
\end{Verbatim}
\end{tcolorbox}

\begin{center}
\includegraphics[width=\linewidth]{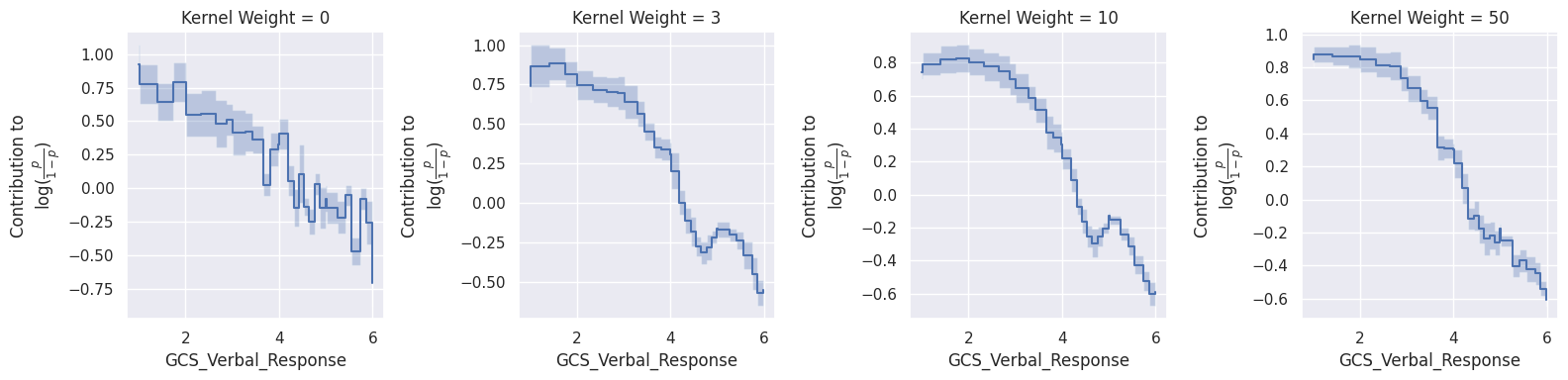}
\end{center}
    
When $\phi = 0$, kernel smoothing is completely
removed, and the shape function is (as expected) very noisy. With $\phi = 3$ and $\phi = 10$, a smoother shape emerges but still with two inflection
points. Finally, when $\phi = 50$, the shape
function becomes nearly monotonic. The default value is 3, and we recommend choosing a value between 1 and 10 for most use cases.

    \hypertarget{monotonic-constraints}{%
\subsubsection{Monotonic Constraints}\label{monotonic-constraints}}

Increasing $\phi$ gives \textbf{dnamite} users one way to
simplify model shape functions. 
Alternatively, users may insist that shape functions are fully
monotonic due to prior domain knowledge of the problem. 
In these cases,
\textbf{dnamite} allows users to provide monotonic constraints on individual features. 
\textbf{dnamite} follows the interpretml API for
user-supplied monotonic constraints: a single list of length
\texttt{n\_features} is required, where 0 indicates no constraints, 1
indicates a monotone increasing constraint, and -1 indicates a monotone
decreasing constraint.
Below we train a \textbf{dnamite} model with selected monotone features by passing the above list to the \texttt{monotone_constraints} optional parameter.

\begin{tcolorbox}[breakable, size=fbox, boxrule=1pt, pad at break*=1mm,colback=cellbackground, colframe=cellborder]
\begin{Verbatim}[commandchars=\\\{\}]
\PY{n}{monotone\PYZus{}constraints} \PY{o}{=} \PY{p}{[}
    \PY{l+m+mi}{1} \PY{k}{if} \PY{n}{c} \PY{o}{==} \PY{l+s+s2}{\PYZdq{}}\PY{l+s+s2}{AGE}\PY{l+s+s2}{\PYZdq{}} \PY{k}{else}
    \PY{o}{\PYZhy{}}\PY{l+m+mi}{1} \PY{k}{if} \PY{n}{c} \PY{o+ow}{in} \PY{p}{[}\PY{l+s+s2}{\PYZdq{}}\PY{l+s+s2}{GCS\PYZus{}Eye\PYZus{}Opening}\PY{l+s+s2}{\PYZdq{}}\PY{p}{,} \PY{l+s+s2}{\PYZdq{}}\PY{l+s+s2}{GCS\PYZus{}Verbal\PYZus{}Response}\PY{l+s+s2}{\PYZdq{}}\PY{p}{]} \PY{k}{else}
    \PY{l+m+mi}{0}
    \PY{k}{for} \PY{n}{c} \PY{o+ow}{in} \PY{n}{X\PYZus{}train\PYZus{}benchmark}\PY{o}{.}\PY{n}{columns}
\PY{p}{]}
\PY{n}{model} \PY{o}{=} \PY{n}{DNAMiteBinaryClassifier}\PY{p}{(}
    \PY{n}{random\PYZus{}state}\PY{o}{=}\PY{n}{seed}\PY{p}{,}
    \PY{n}{n\PYZus{}features}\PY{o}{=}\PY{n}{X\PYZus{}train\PYZus{}benchmark}\PY{o}{.}\PY{n}{shape}\PY{p}{[}\PY{l+m+mi}{1}\PY{p}{]}\PY{p}{,} 
    \PY{n}{device}\PY{o}{=}\PY{n}{device}\PY{p}{,}
    \PY{n}{monotone\PYZus{}constraints}\PY{o}{=}\PY{n}{monotone\PYZus{}constraints}\PY{p}{,}
\PY{p}{)}
\PY{n}{model}\PY{o}{.}\PY{n}{fit}\PY{p}{(}\PY{n}{X\PYZus{}train\PYZus{}benchmark}\PY{p}{,} \PY{n}{y\PYZus{}train\PYZus{}benchmark}\PY{p}{)}
\end{Verbatim}
\end{tcolorbox}

\textbf{dnamite} implements monotonic constraints as follows. Suppose the
input for feature \(j\) falls in bin index \(i\).
Without monotonic constraints, the output for feature \(j\) would be
\(\hat{f}_j(i)\). For a monotone increasing feature, this output is replaced with
\(o_j + \sum_{k \leq i} \hat{f}_j(i - k)^2\), where \(o_j\) is a
learnable offset parameter. Similarly, for a monotone decreasing
feature, the output is replaced with
\(o_j - \sum_{k \leq i} \hat{f}_j(i - k)^2\). 

Below we compare the AUC of this model to the model trained in \ref{basic-usage}, which has no monotonicity constraints.

    \begin{tcolorbox}[breakable, size=fbox, boxrule=1pt, pad at break*=1mm,colback=cellbackground, colframe=cellborder]
\begin{Verbatim}[commandchars=\\\{\}]
\PY{n}{preds} \PY{o}{=} \PY{n}{model}\PY{o}{.}\PY{n}{predict\PYZus{}proba}\PY{p}{(}\PY{n}{X\PYZus{}test\PYZus{}benchmark}\PY{p}{)}
\PY{n}{monotone\PYZus{}constraints\PYZus{}auc} \PY{o}{=} \PY{n}{roc\PYZus{}auc\PYZus{}score}\PY{p}{(}\PY{n}{y\PYZus{}test\PYZus{}benchmark}\PY{p}{,} \PY{n}{preds}\PY{p}{)}
\PY{n+nb}{print}\PY{p}{(}\PY{l+s+sa}{f}\PY{l+s+s2}{\PYZdq{}}\PY{l+s+s2}{Monotone constraints AUC: }\PY{l+s+si}{\PYZob{}}\PY{n+nb}{round}\PY{p}{(}\PY{n}{monotone\PYZus{}constraints\PYZus{}auc}\PY{p}{,}\PY{+w}{ }\PY{l+m+mi}{4}\PY{p}{)}\PY{l+s+si}{\PYZcb{}}\PY{l+s+s2}{\PYZdq{}}\PY{p}{)}
\PY{n+nb}{print}\PY{p}{(}\PY{l+s+sa}{f}\PY{l+s+s2}{\PYZdq{}}\PY{l+s+s2}{Baseline AUC: }\PY{l+s+si}{\PYZob{}}\PY{n}{dnamite\PYZus{}auc}\PY{l+s+si}{\PYZcb{}}\PY{l+s+s2}{\PYZdq{}}\PY{p}{)}
\end{Verbatim}
\end{tcolorbox}

    \begin{Verbatim}[commandchars=\\\{\}]
Monotone constraints AUC: 0.8484
Baseline AUC: 0.8521
    \end{Verbatim}

The monotonicity constraints have little effect on the predictive performance.
Further, we can plot shape functions to verify that the monotonic constraints are working as intended.

\begin{tcolorbox}[breakable, size=fbox, boxrule=1pt, pad at break*=1mm,colback=cellbackground, colframe=cellborder]
\begin{Verbatim}[commandchars=\\\{\}]
\PY{n}{model}\PY{o}{.}\PY{n}{plot\PYZus{}shape\PYZus{}function}\PY{p}{(}\PY{p}{[}
    \PY{l+s+s2}{\PYZdq{}}\PY{l+s+s2}{AGE}\PY{l+s+s2}{\PYZdq{}}\PY{p}{,}
    \PY{l+s+s2}{\PYZdq{}}\PY{l+s+s2}{Respiratory\PYZus{}Rate}\PY{l+s+s2}{\PYZdq{}}\PY{p}{,}
    \PY{l+s+s2}{\PYZdq{}}\PY{l+s+s2}{GCS\PYZus{}Eye\PYZus{}Opening}\PY{l+s+s2}{\PYZdq{}}\PY{p}{,}
    \PY{l+s+s2}{\PYZdq{}}\PY{l+s+s2}{GCS\PYZus{}Verbal\PYZus{}Response}\PY{l+s+s2}{\PYZdq{}}\PY{p}{,}
\PY{p}{]}\PY{p}{)}
\end{Verbatim}
\end{tcolorbox}

\begin{center}
\includegraphics[width=\linewidth]{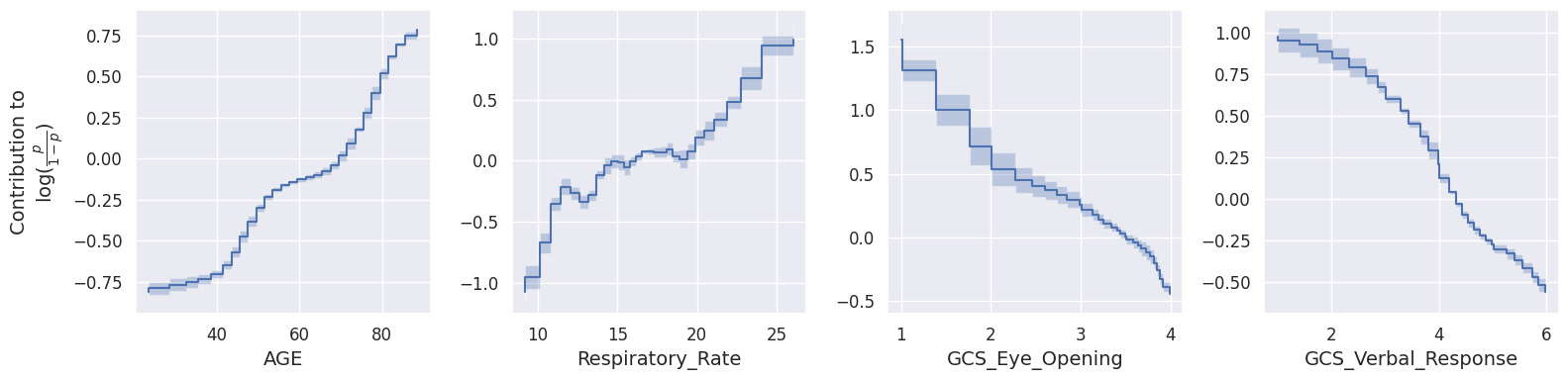}
\end{center}

The shape functions for age, GCS eye opening, and GCS verbal response follow their intended monotonicity constraints, while the respiratory rate shape function is not monotonic.

    \hypertarget{feature-selection-paths}{%
\subsubsection{Feature Selection Paths}\label{feature-selection-paths}}

Determining the right number of features to choose can be challenging in feature selection models.
\textbf{dnamite} controls the number of features chosen through \texttt{reg_param} ($\lambda$ in Equation \ref{eq:ss_loss}). 
One strategy sets $\lambda$ via hyperparameter optimization using validation predictive performance. 
However, this strategy will prioritize accuracy over interpretability, 
producing models with (possibly too) many features.

An alternative approach is to test multiple 
$\lambda$ values and select the smallest feature set that maintains competitive predictive performance.
To facilitate this approach, \textbf{dnamite} implements a function
\texttt{get\_regularization\_path} that helps users select the right
number of features for their dataset. 
The function tries several
$\lambda$ values and reports the
validation loss and score associated with each value.

    \begin{tcolorbox}[breakable, size=fbox, boxrule=1pt, pad at break*=1mm,colback=cellbackground, colframe=cellborder]
\begin{Verbatim}[commandchars=\\\{\}]
\PY{n}{model} \PY{o}{=} \PY{n}{DNAMiteBinaryClassifier}\PY{p}{(}
   \PY{n}{random\PYZus{}state}\PY{o}{=}\PY{n}{seed}\PY{p}{,}
   \PY{n}{n\PYZus{}features}\PY{o}{=}\PY{n}{X\PYZus{}train\PYZus{}cls}\PY{o}{.}\PY{n}{shape}\PY{p}{[}\PY{l+m+mi}{1}\PY{p}{]}\PY{p}{,} 
   \PY{n}{device}\PY{o}{=}\PY{n}{device}\PY{p}{,}
\PY{p}{)}
\PY{n}{feats}\PY{p}{,} \PY{n}{path\PYZus{}data} \PY{o}{=} \PY{n}{model}\PY{o}{.}\PY{n}{get\PYZus{}regularization\PYZus{}path}\PY{p}{(}
   \PY{n}{X\PYZus{}train\PYZus{}cls}\PY{p}{,} \PY{n}{y\PYZus{}train\PYZus{}cls}\PY{p}{,} \PY{n}{init\PYZus{}reg\PYZus{}param}\PY{o}{=}\PY{l+m+mf}{0.01}
\PY{p}{)}
\PY{n}{plt}\PY{o}{.}\PY{n}{plot}\PY{p}{(}
   \PY{n}{path\PYZus{}data}\PY{p}{[}\PY{l+s+s2}{\PYZdq{}}\PY{l+s+s2}{num\PYZus{}feats}\PY{l+s+s2}{\PYZdq{}}\PY{p}{]}\PY{p}{,} \PY{n}{path\PYZus{}data}\PY{p}{[}\PY{l+s+s2}{\PYZdq{}}\PY{l+s+s2}{val\PYZus{}loss}\PY{l+s+s2}{\PYZdq{}}\PY{p}{]}\PY{p}{,} 
   \PY{n}{marker}\PY{o}{=}\PY{l+s+s1}{\PYZsq{}}\PY{l+s+s1}{o}\PY{l+s+s1}{\PYZsq{}}\PY{p}{,} \PY{n}{color}\PY{o}{=}\PY{l+s+s1}{\PYZsq{}}\PY{l+s+s1}{b}\PY{l+s+s1}{\PYZsq{}}
\PY{p}{)}
\PY{n}{plt}\PY{o}{.}\PY{n}{plot}\PY{p}{(}
   \PY{n}{path\PYZus{}data}\PY{p}{[}\PY{l+s+s2}{\PYZdq{}}\PY{l+s+s2}{num\PYZus{}feats}\PY{l+s+s2}{\PYZdq{}}\PY{p}{]}\PY{p}{,} \PY{n}{path\PYZus{}data}\PY{p}{[}\PY{l+s+s2}{\PYZdq{}}\PY{l+s+s2}{val\PYZus{}AUC}\PY{l+s+s2}{\PYZdq{}}\PY{p}{]}\PY{p}{,} 
   \PY{n}{marker}\PY{o}{=}\PY{l+s+s1}{\PYZsq{}}\PY{l+s+s1}{o}\PY{l+s+s1}{\PYZsq{}}\PY{p}{,} \PY{n}{color}\PY{o}{=}\PY{l+s+s1}{\PYZsq{}}\PY{l+s+s1}{r}\PY{l+s+s1}{\PYZsq{}}
\PY{p}{)}
\PY{n}{plt}\PY{o}{.}\PY{n}{legend}\PY{p}{(}\PY{p}{[}\PY{l+s+s2}{\PYZdq{}}\PY{l+s+s2}{Validation Loss}\PY{l+s+s2}{\PYZdq{}}\PY{p}{,} \PY{l+s+s2}{\PYZdq{}}\PY{l+s+s2}{Validation AUC}\PY{l+s+s2}{\PYZdq{}}\PY{p}{]}\PY{p}{)}
\end{Verbatim}
\end{tcolorbox}

            \begin{tcolorbox}[breakable, size=fbox, boxrule=.5pt, pad at break*=1mm, opacityfill=0]
\begin{Verbatim}[commandchars=\\\{\}]
<matplotlib.legend.Legend at 0x7f80dbec7e50>
\end{Verbatim}
\end{tcolorbox}
    
\begin{center}
\includegraphics[width=0.6\linewidth]{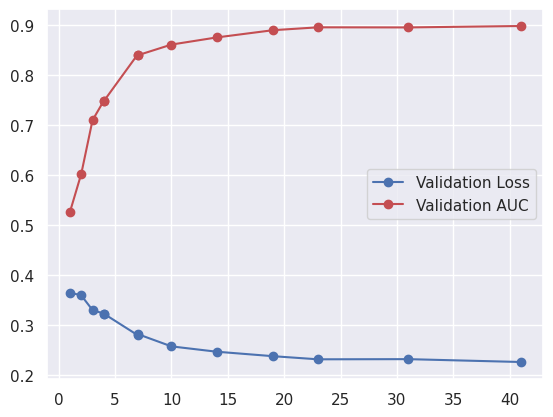}
\end{center}
    
Selecting less than 10 features yields
significantly worse predictive performance, while the performance
improvements after selecting about 15 features are marginal. To
see the features that are selected at each point in this graph, the
returned dictionary \texttt{feats} can be queried using the number of
selected features as the key.

    \begin{tcolorbox}[breakable, size=fbox, boxrule=1pt, pad at break*=1mm,colback=cellbackground, colframe=cellborder]
\begin{Verbatim}[commandchars=\\\{\}]
\PY{n}{feats}\PY{p}{[}\PY{l+m+mi}{10}\PY{p}{]}
\end{Verbatim}
\end{tcolorbox}

\begin{Verbatim}[commandchars=\\\{\}]
['AGE',
 'lab\_Lactate\_mmol/L',
 'lab\_Anion\_Gap\_mEq/L',
 'lab\_Bicarbonate\_mEq/L',
 'lab\_Sodium\_mEq/L',
 'lab\_RDW\_\%',
 'chart\_Level\_of\_Conscious',
 'chart\_Activity',
 'chart\_Code\_Status',
 'chart\_Orientation']
\end{Verbatim}

\section{Conclusion}
\label{sec:conclusion}

This paper introduces \textbf{dnamite}, a Python package that implements a suite of Neural Additive Models (NAMs).
\textbf{dnamite}'s models can be used for regression, classification, and survival analysis, offering improved accuracy compared to linear models and improved interpretability compared to black-box models.
Further, \textbf{dnamite} supports automated feature selection before training, allowing users to fit low-dimensional additive models on high-dimensional datasets.
By using the DNAMite architecture, \textbf{dnamite}'s models can flexibly adapt to different desired smoothness levels in shape functions, as well as elegantly handle categorical features and missing values.
Our usage demonstration on the MIMIC III dataset illustrates the power of \textbf{dnamite} as a package for interpretable machine learning.
We hope that \textbf{dnamite} will inspire more work making NAMs and other glass-box machine learning models more useful in practice.

\bibliography{bib}

\end{document}